\documentclass[journal]{IEEEtran}
\IEEEoverridecommandlockouts
\ifCLASSINFOpdf
  \usepackage[pdftex]{graphicx}
\else
\fi
\usepackage{xcolor}

\usepackage{amsmath}
\usepackage{amssymb}

\begin{document}
%
\title{Maximising Wrenches for Kinematically Redundant Systems with Experiments on UVMS}

\author{Wilhelm J. Marais$^{1}$, Stefan B. Williams$^{1}$, Oscar Pizarro$^{1}$
\thanks{$^{1}$Authors are with Australian Centre for Field Robotics, University of Sydney, NSW Australia,
        {\tt\footnotesize (j.marais, stefanw, o.pizarro)@acfr.usyd.edu.au}}
}


%


\maketitle

\begin{abstract}
This paper presents methods for finding optimal configurations and actuator forces/torques to maximise contact wrenches in a desired direction for Underwater Vehicles Manipulator Systems (UVMS). The wrench maximisation problem is formulated as a linear programming problem, and the optimal configuration is solved as a bi-level optimisation in the parameterised redundancy space. We additionally consider the cases of one or more manipulators with multiple contact forces, maximising wrench capability while tracking a trajectory, and generating large wrench impulses using dynamic motions. We look at the specific cases of maximising force to lift a heavy load, and maximising torque during a valve turning operation. Extensive experimental results are presented using an underwater robotic platform equipped with a 4DOF manipulator, and show significant increases in wrench capability compared to existing methods for UVMS.
\end{abstract}

\section{Introduction}
%
%
%
%
High Degree Of Freedom (DOF) vehicle-manipulator systems have seen increased use in both industrial and field robotics settings due to the advantages provided by kinematic redundancy. Since these systems have a large number of degrees of freedom responsible for end effector motion, kinematic positioning generally requires iterative techniques. Despite this, these kinematically redundant systems can make use of the continuous space of configurations which solve a particular inverse kinematics problem to optimise additional secondary objectives~\cite{reviewPseudoinverse}. Secondary objectives may include obstacle avoidance, optimisation of dynamic manipulability and avoidance of configurations limits~\cite{AutonomousPanel, underwaterRobots}. 
During interaction tasks with objects or the environment, end effector poses as well as wrenches are of consideration~\cite{unifiedForceControl, forceControlUVMS}. External contact forces and torques between the end effector and the environment are transmitted through each DOF of the system responsible for end effector motion, in a way which is highly dependent on the configuration. Exploiting the continuous redundancy offered by high DOF systems allows for configurations which simultaneously achieve a desired end effector pose, and maximise the wrench capabilities~\cite{pragmaticFullForce}.

Interaction tasks in underwater environments for scientific exploration or industrial purposes have been traditionally conducted by medium to large Underwater Vehicle Manipulator Systems (UVMS), requiring specialised equipment and multiple operators for launch and recovery as well as operation~\cite{underwaterManipReview}.
In recent years, the emergence of smaller and lower cost commercial off-the-shelf underwater vehicles and manipulators has seen a transition from larger to small vehicles for some interaction tasks~\cite{coupledLightWeightUVMS, adpativeControlUVMS}. This reduces both operations costs and time, as well as increasing the accessibility of these systems since they can be launched, operated, and recovered by a small team with no specialised equipment. 
This work looks at maximising the capability of these small systems, allowing them to perform a larger range of interaction tasks. Specifically, maximising contact wrenches between the end effector and the environment are considered. This work assumes the object to be manipulated has already been firmly grasped by the end effector of the UVMS, and the system can then transition to a configuration which allows for the maximum wrench to be applied at the end effector.

\begin{figure} [t]
   \centering
   \includegraphics[width=1.0\linewidth]{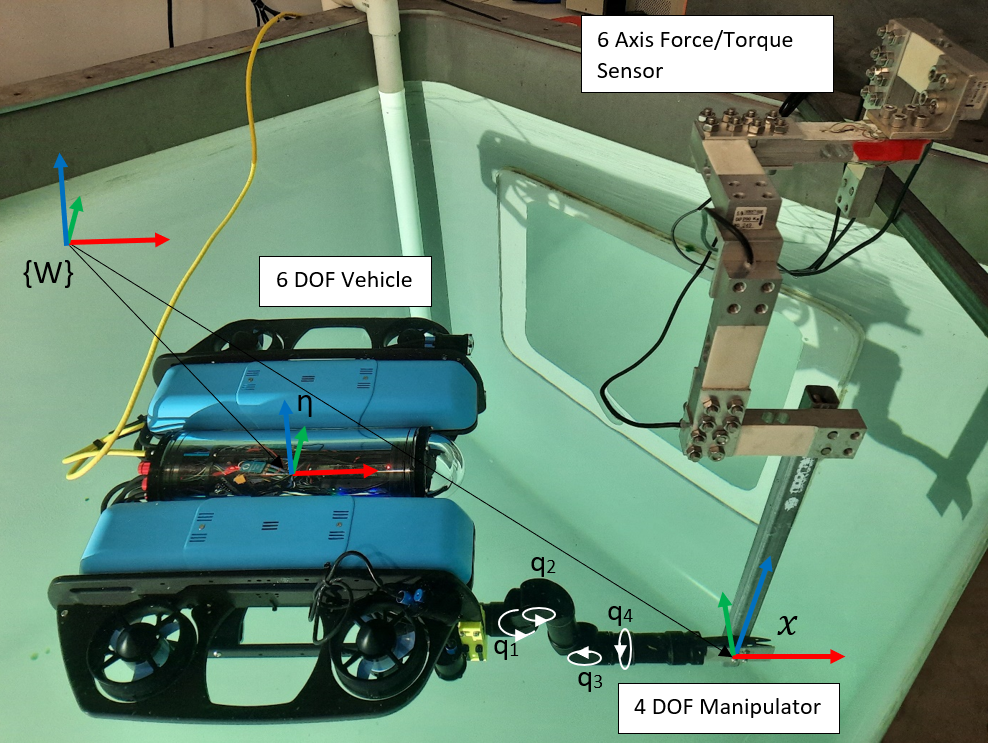}
   \caption{UVMS used in experiments with a 6 DOF Bluerov vehicle and 4 DOF Blueprint Lab Reach Alpha manipulator, showing vehicle pose $\eta$ and end effector pose $x$ relative to the world frame $\{ W \}$. Manipulator joints are labelled $q_1$ to $q_4$}
   \label{fig:simulation}
\end{figure}

We compare a number of cases in which wrench capability should be maximised. These include the static case, where the aim is to maximise the wrench at a single configuration, which we further extended to consider multiple contact points with the environment. The second case considers a trajectory where a given desired end effector path should be tracked, while finding a set of configurations along the way which are dynamically feasible and maximise the lowest applicable wrench along the path. Finally, the case of generating large wrench impulses for a fixed end effector pose is considered. Extensive experimental validation of the proposed methods is provided, which shows significant increases in wrench capability compared to previous methods for UVMS.

The contributions of this paper are the following:
\begin{itemize}
    \item A bi-level optimisation approach for finding optimal configurations and actuator efforts for maximising wrenches for UVMS is proposed. Experimental results show this provides significantly larger wrenches than existing transmission ratio optimisation methods
    \item Consideration of relaxing constraints on orthogonal wrenches for certain tasks, with experimental results showing a threefold increase in maximum wrench capability
    \item An optimisation method is proposed for maximising wrenches for UVMS with multiple contact points, including parameterisation of a set of secondary grasping points, and analysis of the required constraints. Experimental results show increased wrench capability using the proposed methods
    \item A bi-level optimisation method is proposed for generating trajectories which maximise wrenches over a trajectory, with experimental results confirming the validity of the proposed method
    \item A proposed method for generating heaving motions for large wrench impulses, with experimental results showing the validity of the approach.
\end{itemize}

The rest of this paper is structured as follows. Section~\ref{sec: Related Work} gives a recap of the mathematical background and current methods for wrench analysis and finding optimal configurations for wrench maximisation. Section~\ref{sec: Static Wrench} describes the methods proposed in this work for maximising static wrench capability for UVMS, including consideration of multiple contact points. Section~\ref{sec: Trajectory} extends the methods used for static analysis to consider dynamic trajectories given an end effector path. Section~\ref{sec: Impulse} describes the proposed method for generating large wrench impulses using dynamic motions. Finally, Section~\ref{sec: Results} presents experimental results for each section, followed by concluding remarks and directions for future work in Section~\ref{sec: Conclusions}.

\section{Background and Related Work}
\label{sec: Related Work}

\subsection{Kinematic and Dynamic Modelling}
A vehicle manipulator system  has system configuration $\theta = (\eta,q)^T$, with $dim(\theta) = n$, where $\eta \in SE(3)$ is the vehicle pose in the world frame, and $q \in \mathbb{R}^{n-6}$ is the manipulator joint angles. These are shown in Figure~\ref{fig:simulation}. Given some end effector pose $x \in SE(3)$, in the world frame, the forward non-linear map $f_k$ gives
\begin{equation}
    x = f_k(\theta)
\end{equation}
for which analytical solutions in the reverse direction, the inverse kinematics problem, are generally not available. Numerical solutions make use of the linear velocity relationship
\begin{equation}
    \dot{x} = J(\theta) \dot{\theta}
\end{equation}
where $\dot{x} \in \mathbb{R}^6$ is the end effector velocity vector in the inertial frame, $\dot{\theta} \in \mathbb{R}^n$ is the vector of system velocities corresponding to each DOF, and $J(\theta)$ is the configuration dependant Jacobian. The $\theta$ dependence for $J$ is dropped in further notation. 
For kinematically redundant systems this has a least squares solution for a desired $\dot{x}$ given by
\begin{equation}
    \dot{\theta} = J^+\dot{x} 
\end{equation}
where $J^+$ is the pseudoinverse of $J$. Note in this work we consider fully actuated vehicles, therefore the Jacobian for a single manipulator system always has full row rank. 
The null space of $J$, defined as the set of system velocities which cause no end effector velocity can be included as
\begin{equation}
    \dot{\theta} = J^+\dot{x} + \alpha(I - J^+J)\nabla H 
    \label{eq: NullSpace}
\end{equation}
where $\alpha$ is a scaling factor, and $H$ is some secondary objective to be optimised in the null space~\cite{reviewPseudoinverse}. The null space projection $(I - J^+J)$ can be thought of as a set of (possibly non linearly independent) basis vectors which is tangent to the inverse kinematics equality constraint. 

Now we have the end effector wrench vector 
\begin{equation}
    h_e = \left( \begin{array}{c}
        ^0f_e   \\
        ^0n_e  
    \end{array} \right) \in \mathbb{R}^6
\end{equation}
consisting of end effector forces $^0f_e$ and torques $^0n_e$ in the inertial frame, with force relationship 
\begin{equation}
    \tau_h = J^T h_e
    \label{eq: endToTau}
\end{equation}
where $\tau_h \in \mathbb{R}^n$ is the vector of forces and torques on each DOF due to $h_e$.
The actuator model is given by $Bu$, where $u \in [u_{min},u_{max}]$ in $\mathbb{R}^m$ is the control input vector for each actuator, and the matrix $B \in \mathbb{R}^{n \times m}$ is the mapping between actuator control input force/torque on each DOF. 
Including the dynamics of the systems gives
\begin{equation}
    M(\theta)\ddot{\theta} + C(\theta,\dot{\theta})\dot{\theta}  + D(\theta,\dot{\theta})\dot{\theta} + g(\theta) + J^T h_e = Bu  
    \label{eq:dynamics}
\end{equation}
where $M(\theta)$ is the configuration dependent mass matrix, $C(\theta,\dot{\theta})$ is the Coriolis terms, $D(\theta,\dot{\theta})$ is damping terms modelled using a combination of linear and quadratic drags, and $g(\theta)$ is the vector of gravity and buoyancy forces. We collect all dynamic terms on the the left into the vector $\tau_d$, giving
\begin{equation}
   \tau_d + J^T h_e = Bu 
   \label{eq:reducedDynamics}
\end{equation}

\subsection{Ellipsoids and Polytopes}
Analysis of wrench capability typically involves ellipsoids and polytopes.
 Velocity ellipsoids and their counterparts, wrench ellipsoids, were introduced by \cite{yoshikawaManipulability}. These provide a mapping from a unit ball of joint torques to end effector wrench, defined as

\begin{equation}
    \{ h_e \; \mid \; \|J^T h_e \|_2 \leq 1 \}
\end{equation}
Recent work proposed maximising the volume of the velocity ellipsoid by projecting the gradient onto the null space of the system as in Equation~\ref{eq: NullSpace}, and further by solving the problem as a quadratic program~\cite{QPManipulability} . A similar null space projection method was used by \cite{optimalValveTurning} to maximise the dynamic manipulability ellipsoid during a value turning operation for an UVMS. 

The volume of the wrench ellipsoid accounts for the ability of the system to apply forces and torques in all directions in an isotropic manner. In this work, it is desired to maximise the wrench along a single force/torque direction, therefore anisotropic capability measures are more appropriate. 
One such measure is the transmission ratio, which is defined as the radius of the wrench ellipsoid in a desired direction, and can been optimised by null space projection~\cite{transmissionRatio}.
To find the distance to the wrench ellipsoid in a given direction, consider a unit vector $c \in \mathbb{R}^6$ which defines the wrench direction, and scalar $\beta$, with the condition that it lies on the ellipse given by
\begin{equation}
    (\beta c)^T JJ^T (\beta c) = 1
\end{equation}
and solving for $\beta$,
\begin{equation}
    \beta = (c^T JJ^T c)^{-1/2}
\end{equation}
This analysis does not include the effect of gravity, buoyancy and dynamical effects which offsets the centre of the ellipsoid, giving a dynamic wrench ellipsoid defined as
\begin{equation}
    \{ h_e \; \mid \; \|J^T h_e + \tau_d\|_2 \leq 1 \}
\end{equation}

For an ellipse with centre not at the origin, we can solve for $\beta$ assuming the origin is contained in the ellipsoid. This condition is met with a system which can support its own weight under gravity and sustain the dynamic loads. We solve
\begin{equation}
    (J^T\beta c +  \tau_d)^T(J^T\beta c + \tau_d) = 1
\end{equation}
for $\beta$ by taking the positive root.

Symmetric torque limits have been proposed to  be incorporated into the transmission ratio~\cite{enhancementForceExertion}, yet dynamical effects were neglected. 
Some works have considered minimising the sum of squared torques on each joint for a given wrench of a redundant serial manipulator by null space projection~\cite{forceControlRedundant}, which is functionally equivalent to optimising the transmission ratio.

The idea of wrench ellipsoids was extended by~\cite{choosingStiffness} by scaling the ellipsoid by a desired force/stiffness matrix, and instead optimising the trace of the resultant ellipsoid (the sum of the square of the radii), to get a configuration dependant measure. Further work~\cite{dynamicCapabilityEquations} looked at combining velocity, acceleration and wrench capabilities in the joint torque space for serial manipulators. This involves mapping maximal balls of velocity, acceleration and wrench individually onto ellipsoids in the joint torque space, and then combining to form a torque hypersurface which has to satisfy actuator constraints. Since the method combines worst-case scenarios for velocities, accelerations and wrenches, it provides a very conservative measure of manipulator capability.

It has been shown that wrench ellipsoids provide only an approximate measure of manipulator performance as it fails to capture to the true constraints of the system, as compared to wrench polytopes which provides a better description of the true capabilities of a system~\cite{wrenchPolytope}. By finding the set of effector wrenches which satisfy joint constraints, the wrench polytope is defined as
\begin{equation}
    \{ h_e \; \mid \; \tau_{min} \leq (J^T h_e + \tau_d) \leq \tau_{max} \}
\end{equation}
where $\tau_{min}$ and $\tau_{max}$ are the minimum and maximum loads on each DOF. 
Wrench polytope analysis has been used extensively in the design and evaluation of manipulators and manipulator poses \cite{wrenchParallelPart1, wrenchCapabilitiesPolytope}, yet optimisation is difficult since the quality of a polytope is difficult to quantify. Some attempts to define capability measures using polytopes involve computing actuator saturation for a given wrench direction for serial manipulators~\cite{pragmaticFullForce}, or solving Linear Programming (LP) problems for parallel manipulators~\cite{hapticRedundancy, ParallelMaximumWrench}.

\subsection{Redundancy Parameterisation}
The idea of redundancy parameterisation has been proposed in several works to greatly reduce the number of dimensions over which to optimise a given performance objective, and to remove the non-linear inverse kinematics constraint. This methods avoids the use of the Jacobian null space projection in Equation~\ref{eq: NullSpace} and instead explicitly considers a minimal set of DOF which describe the available self motions while keeping a fixed end effector pose. Redundancy parameterisation has been used for choosing optimal stiffness configurations in serial manipulators with one degree of redundancy ~\cite{bussonRigidity}, as well as for exploring the force capabilities of a manipulator to apply forces normal to a surface~\cite{optimalContactForce}. This work was extended in \cite{pragmaticFullForce} to an exhaustive search over 2 redundant DOF for a 7DOF manipulator, where the saturating wrench was used to find the weight lifting capability for each pose, while \cite{forceOptimisationPlanar} examined a planar parallel manipulator with a 3DOF redundant space to minimise the sum of squared joint torques for a given wrench.
Given an appropriate parameterisation of the redundant DOF written as  $\theta_r \in \mathbb{R}^{n-6}$, Equation~\ref{eq: NullSpace} can be re-written as 
\begin{equation}
    \dot{\theta} = J_e\dot{x} + A_r \dot{\theta_r} 
    \label{eq: redundantJacobian}
\end{equation}
where $\dot{\theta_r} \in \mathbb{R}^{n-6}$ is the vector of velocities of the redundant DOF, $A_r \in \mathbb{R}^{n \times (n-6)}$ is the null space projection matrix which maps redundant velocities to system velocities, and $J_e \in \mathbb{R}^{n \times 6}$ is the Jacobian which maps end effector velocities to all DOF which are not redundant DOF.

\subsection{Wrench Capability Over a Trajectory}
The previous works have considered wrench capabilities for a single end effector pose. In some cases the wrench capability over an entire trajectory has to be considered. Local redundancy resolution methods have been used for continuous wrench maximisation over a trajectory~\cite{optimalValveTurning}, yet local methods may lead to numerical instability and sub-optimal trajectories~\cite{globalvsLocal}. Global redundancy resolution methods which consider entire paths have been proposed for trajectory planning in kinematically redundant systems~\cite{localvsglobalTorque}.
Early works in global redundancy resolution~\cite{largeForcePlanning, manipulatorLargeForce} looked at planning paths which avoid actuator saturation while experiencing a large fixed load. This has been extended to consider maximising the allowed load~\cite{}, and incorporation of dynamical effects~\cite{}. \cite{ParallelMaximumWrench} used an evolutionary algorithm combined with multiple lower level linear programming solutions to find the maximum wrench capability over a trajectory for a re-configurable parallel robot. 

Hamiltonian control approaches which make use of Pontrayagin's minimum principle over an entire trajectory~\cite{maximumPayloadMobileManipualtor, maximumLoadNonholonomic} have been used for finding maximum load carrying capacity for mobile manipulators, yet often lead to switching control laws which are not physically realisable. Additionally, these methods require solving two point boundary value problems and therefore are generally restricted to systems with low degrees of redundancy. Recent work has considered receding horizon style planning using Model Predictive Control (MPC) to track desired wrenches~\cite{MPCAdmittanceControl}, yet little work has been done to apply this technique to maximising wrench capability. 

\section{Maximising Static Wrench Capability}
\label{sec: Static Wrench}
This section details how to find optimal configurations and actuator efforts to maximise the wrench capability in a desired direction, for a single end effector pose and system configuration. Dynamic effects due to velocity and acceleration are therefore ignored, yet are considered in Section~\ref{sec: Trajectory}. Examples where static wrench capability is desired is for maximising weight lifting capability  or torque during a valve turning operation during a relatively slow operations. 

\begin{figure} [t]
\centering
   \includegraphics[width=1.0\linewidth]{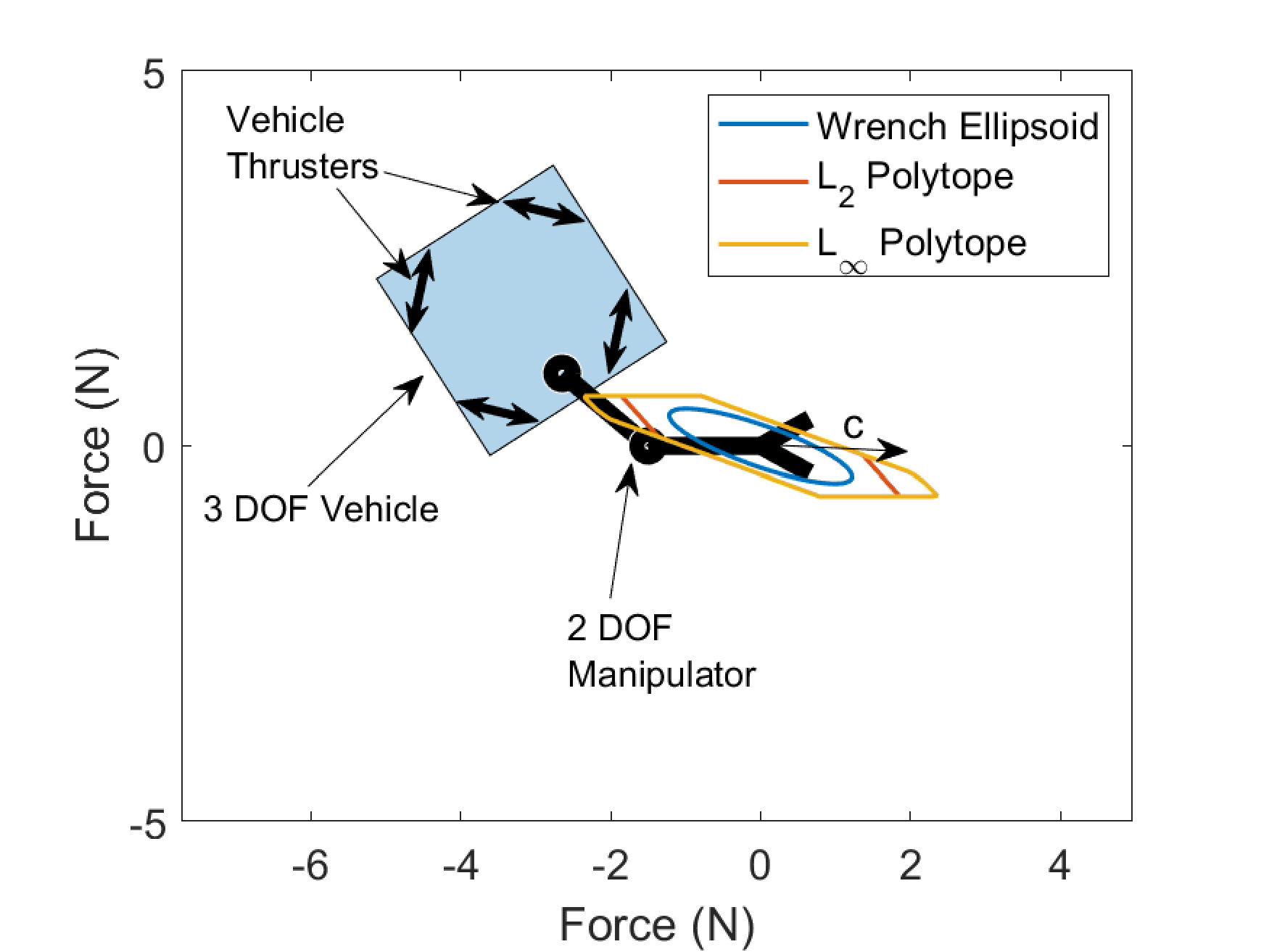}
   \caption{Wrench ellipsoid and polytopes for a 2D UVMS showing force capability in the given configuration and the desired direction $c$ in which the wrench capability should be maximised}
   \label{fig:polytopes}
\end{figure}

First, the previous definitions of wrench ellipsoids and polytopes have to be extended to include the overactuated mapping given in equation~\ref{eq:dynamics}. We can invert the mapping in equation~\ref{eq:reducedDynamics} to get
\begin{equation}
 u = B^+(\tau_d + J^Th_e)
\end{equation}
giving a new definition for the wrench ellipsoid
\begin{equation}
    \{ h_e \; \mid \; \|W_u B^+(J^T h_e + \tau_d)\|_2 \leq 1 \}
\end{equation}
where the actuator weighting matrix~\cite{enhancementForceExertion} is given by $W_u = diag[\frac{1}{u_{max_1}}, ... ,\frac{1}{u_{max_m}}]$ where $u_{max_i}$ is the positive force/torque limit for actuator $i$. In the case of asymmetric limits as in the case of an underwater thruster, the smallest absolute value is used. 
The transmission ratio is also redefined as the positive root of 
\begin{equation}
    (W_u B^+(J^T h_e + \tau_d))^T(W_u B^+(J^T h_e + \tau_d)) = 1
    \label{eq:transmissionUVMS}
\end{equation}
for a given desired wrench along direction $c$. 
Using the inverted actuator mapping to redefine the wrench polytope gives

\begin{equation}
    \{ h_e \; \mid \; u_{min} \leq B^+ (J^T h_e + \tau_d) \leq u_{max} \}
\end{equation} 
although $B^+$, the pseudoinverse of $B$, does not give a true measure of the capabilities of the system. Thus this is referred to as the $L_2$ polytope, while the $L_{\infty}$ polytope is given by 
\begin{equation}
    \{ h_e \; \mid \; (Bu)_{min} \leq J^T h_e + \tau_d \leq (Bu)_{max} \}
\end{equation}
which gives a true measure of the actuator capabilities.
A simple 2-Dimensional (2D) test case is considered to compare each capability measures shown in Figure~\ref{fig:polytopes}. This shows a underwater vehicle with diagonally mounted thrusters on each corner which can apply 1N thrust. The arm has two joints each which can apply a torque of 1Nm. The wrench ellipsoid, $L_2$ and $L_{\infty}$ polytopes along the 2D force dimensions with 0 end effector torque are shown. Clearly the $L_{\infty}$ polytope gives a better measure of the true wrench capabilities of the system.



\begin{figure} [b]
   \centering
   \includegraphics[width=0.8\linewidth]{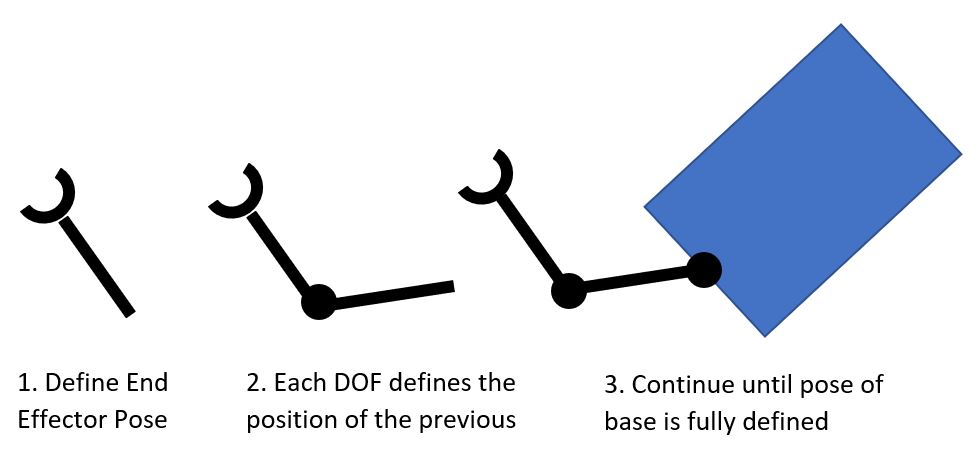}
   \caption{Sequence of diagrams showing method of redundancy parameterisation for UVMS}
   \label{fig:parameterisation}
\end{figure}

Now consider finding the configuration which maximises the wrench capability of the system shown in Figure~\ref{fig:polytopes} in the direction $c$, while keeping the same end effector pose. 
A redundancy parameterisation method for UVMS has been proposed in previous work by the authors~\cite{anisotropicDisturbance}, and is shown in Figure~\ref{fig:parameterisation}. Given a desired end effector pose, each DOF in reverse order starting from the end effector is a redundant DOF until the pose of the base is fully defined, with invalid poses due to self collision discarded. The redundant configuration $\theta_r$, together with the end effector pose $x$ can be used to determine the full system configuration $\theta$. 
For the system considered in Figure~\ref{fig:polytopes}, there are two redundant DOF corresponding to the joints on the manipulator, giving a two dimensional search. Several optimisation objectives can be defined for this optimisation problem. The first is the transmission ratio given as the positive root of Equation~\ref{eq:transmissionUVMS} for the desired wrench direction $c$. This is labelled as optimisation objective $\beta_1$. 
The second is to maximise the $L_{\infty}$ polytope in the direction of $c$. This can be found by solving a linear programming problem given by 
\begin{equation}
    \max_{u,h_e} \quad c^T h_e
\end{equation}
with inequality constraints due to the actuators 
\begin{equation}
   u_{min} \leq u \leq u_{max}
\end{equation}
and equality constraints due to the force/torque balance
\begin{equation}
    \tau_d + J^T h_e = Bu
\end{equation}
Finally there is a constraint on the direction of the applied wrench given by
\begin{equation}
   (cc^+ - I_6)h_e = 0
   \label{eq: orthogonalConstraint}
\end{equation}
where $I_6$ is a $6 \times 6$ identity matrix, which enforces $h_e$ to have no component orthogonal to $c$ in $\mathbb{R}^6$. This is labelled as optimisation objective $\beta_2$. Finally, the orthogonal wrench equality constraint in Equation~\ref{eq: orthogonalConstraint} can be removed in some scenarios. One example is during a valve turning operation, where reaction forces orthogonal to the desired wrench maximisation direction can be applied on the valve. This optimisation objective which relaxes the orthogonality constraint is labelled $\beta_3$. The linear programming problems are solved using the dual-simplex method in Matlab's linprog function.

The aim is to find an optimal configuration which maximises each of the objectives $\beta_1, \beta_2$ and $\beta_3$. This results in a bi-level optimisation framework, where the upper level problem is a search over the parameterised redundancy space to find the configuration which maximises the lower level objective. The lower level objective involves finding the optimal actuator forces/torques to maximise the end effector wrench in the direction of $c$, using the method of either $\beta_1, \beta_2$ or $\beta_3$. 

\begin{figure} [t] 
\centering
   \includegraphics[width=1.0\linewidth]{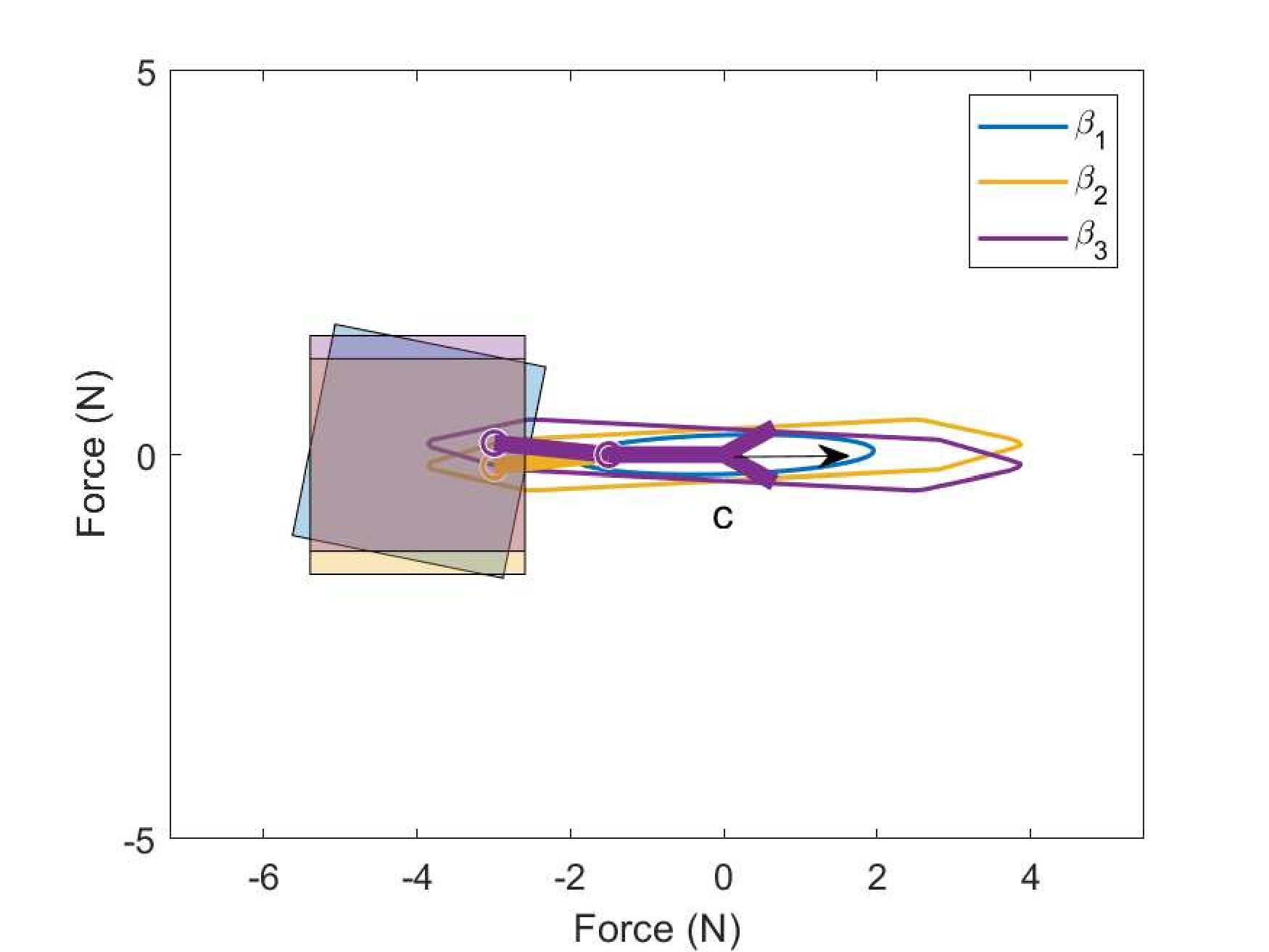}
   \caption{Optimal configurations for a wrench in the $c$ direction for a 2D UVMS, resulting in different optimal configurations according to the measure to be optimised}
   \label{fig:optimalpolytopes}
\end{figure}

Figure~\ref{fig:optimalpolytopes} shows the results of optimising for each of these objectives the same system as in Figure~\ref{fig:polytopes} via a grid search over the 2 redundant DOF, as well as the corresponding ellipsoid or polytopes. The results show that the optimal configuration changes depending on the objective to be optimised. There is also a significant difference in wrench capability. Maximising $\beta_1$ gives a maximum force of $1.91N$, while $\beta_2$ gives $3.47N$ and $\beta_3$ gives $3.88N$. These differences become more pronounced with higher DOF systems, such as the UVMS used for experiments in Section~\ref{sec: Results}. For higher dimensional searches, simulated annealing is used to find a solution with a globally large maximum.

\subsection{Multiple Contact Points}


The case of multiple contact points is common when using two or more manipulators. The reaction forces of the end effector of a second manipulator on some nearby grasp point can help increase the wrench applied by the first manipulator. This creates a closed kinematic chain between the UVMS system and the environment, requiring more careful analysis. 

For multiple contact points case, we define $h_{e_2} \in \mathbb{R}^{6}$, which is the end effector wrench of secondary arm, and $C_2 \in \mathbb{R}^{6 \times l}$, which is a set of $l$ unit vectors which define the direction in which the second manipulator can apply a wrench $h_{e_2}$. This is explained in more detail below. The force/torque balance equation is given by
\begin{equation}
    \tau_d + J^T (h_e, h_{e_2})^T = Bu
\end{equation}
where the terms $t_d, J, B$ and $u$ reflect the additional actuators and DOF of the second manipulator. In order for $(h_e, h_{e_2})^T$ to be well defined, $J \in \mathbb{R}^{12 \times n}$ must have full row rank. This condition is due to the requirement that the system must be able to apply a virtual displacement for a given wrench to be achievable, and is violated at kinematically singular configurations. During searches for optimal configurations, the manipulability measure~\cite{yoshikawaManipulability} is used to discard singular or very near singular configurations. 

The redundancy parameterisation for the dual manipulator case is similar to the single arm case, with the primary arm again defining all degrees of redundancy until the pose of the base is fully defined. The inverse kinematics to reach the grasp point for the second arm can then be analytically solved. In the case of multiple inverse kinematics solutions for the second arm, the solution which maximises the objective is taken. In case of no solution, the configuration is considered invalid. 
A lower level objective can again be written as a linear program
\begin{equation}
    \max_{u,h_e,h_{e_2}} \quad c^T h_e
\end{equation}
subject to 
\begin{equation}
   u_{min} \leq u \leq u_{max}, \; \tau_d + J^T  (h_e, h_{e_2})^T = Bu  
\end{equation}
\begin{equation}
    (cc^+ - I)h_e = 0, \; (C_2C_2^+ - I)h_{e_2} = 0
    \label{eq: dualOrthogonal}
\end{equation}
where as before there is an equality constraint which ensures $h_e$ has no component orthogonal to $c$. The final equality constraint ensures $h_{e_2}$ only has components in allowed directions defined in $C_2$. For example, when firmly grasping a contact point with the secondary manipulator, it might be feasible to set $C_2 = I_6$, meaning any secondary wrench is allowed. If the secondary manipulator is pushing against a flat surface, then $C_2$ may only contain the vector normal to the surface, and the optimisation would include an additional inequality constraint to allow only pushing in one direction.   
\begin{equation}
    C_2 h_{e_2} \geq 0
    \label{eq: inequalityContact}
\end{equation}

The above linear program again is the lower level problem for the bi-level optimisation, this time for finding the optimal configuration which considers multiple contact points and manipulators.
\begin{figure} [t]
   \centering
   \includegraphics[width=0.8\linewidth]{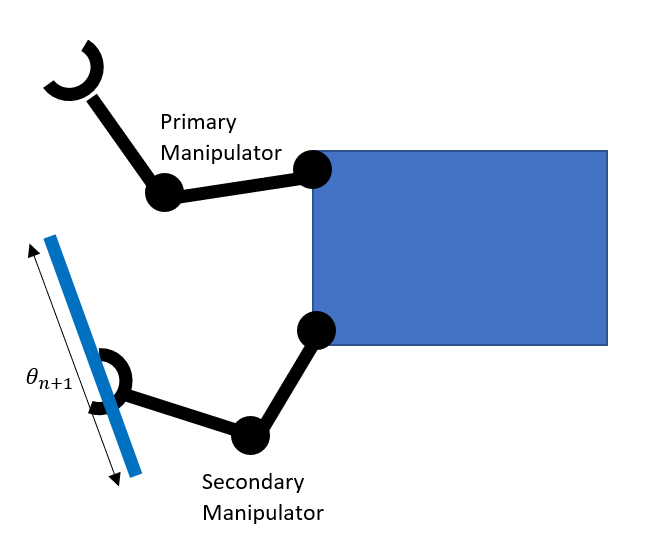}
   \caption{Parametrised regrasp points for second manipulator adding an additional kinematic variable $\theta_{n+1}$ }
   \label{fig:regrasp}
\end{figure}

\begin{figure} [b]
   \centering
   \includegraphics[width = 0.8\columnwidth]{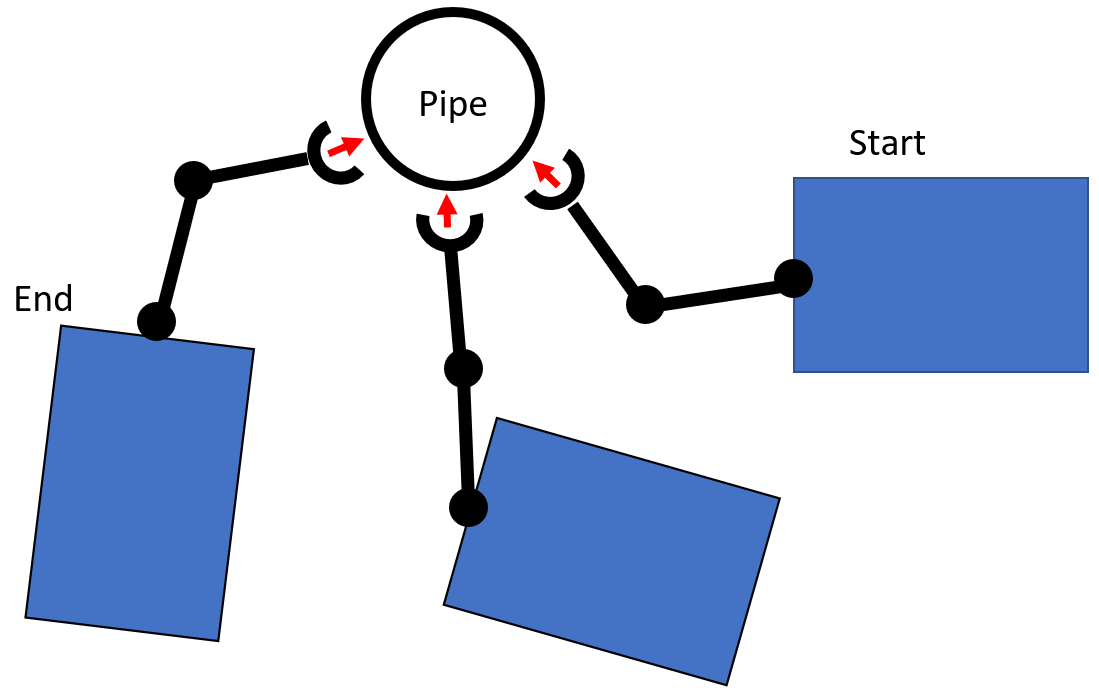}
   \caption{Diagram of vehicle manipulator system showing the starting pose, intermediate poses and ending pose along a rotating end effector trajectory during a pipe waterjet blasting operation. The directions in which the force resisting capability should be maximised for each end effector pose are shown as red arrows.}
   \label{fig:trajectoryDiagram}
\end{figure}
\subsection{Re-grasping}
The above case considered grasping a fixed point with a second arm. Generally multiple grasping points will be available. We consider cases where these regrasp points can be parameterised to provide extra dimensions over which to optimise. Figure~\ref{fig:regrasp} shows a simple 2D example, where a one dimensional set of parameterised re-grasping points result in an additional kinematic degree of freedom which can be optimised over. Section~\ref{sec: Results} describes another parametrisation of re-grasping points.

\section{Wrench Maximisation over a Trajectory}
\label{sec: Trajectory}

In this section, a predefined end effector trajectory is considered, with corresponding desired wrenches along the path. Only a UVMS with a single manipulator is considered, yet the analysis can be easily extended to multiple manipulators.
Use cases for this method may be during a valve turning operation which requires maximising torque along the same direction throughout, or a waterjet blasting operation around a pipe which requires resisting large forces normal to the pipe surface along the path, shown in Figure~\ref{fig:trajectoryDiagram}. Dynamical effects due to velocities and accelerations can not be ignored in this case.
It is assumed there is a given fully defined end effector pose trajectory parameterised in time, and that the velocities at the start and end are zero. To maximise the capability of the system along this path, the objective is to maximise the minimum wrench along this trajectory.

\subsection{Problem Formulation}
In order to make the problem tractable, the trajectory of the end effector is discretised into a set of $k$ successive poses $x_{1},...x_{k}$, written as stacked vector $\boldsymbol{x}$, with corresponding redundant configurations $\theta_{r,1},...\theta_{r,k}$ written as $\boldsymbol{\theta_r}$, giving corresponding system configurations $\theta_{1},...\theta_{k}$, written as $\boldsymbol{\theta}$. Each point along the trajectory of total time $T$ is equally separated by $\Delta t = T/k$. At each end effector pose is a corresponding end effector velocity $\dot{x}_{1},...\dot{x}_{k}$ written as $\boldsymbol{\dot{\theta}}$, which is computed using the timestep and differences between successive poses in $SE(3)$.
Dynamic quantities are computed using the finite difference operator
\begin{equation}
     \mathcal{D}_i = \frac{1}{\Delta t} \left( \begin{array}{ccccc}
        -1  & 1 &   & &     \\
          & -1 & 1 & &     \\
           &  &  \ddots & \ddots  &    \\
           &   & & -1  & 1   \\
          &  &  & & 0   
    \end{array} \right) \otimes I_{i \times i}
\end{equation}
where $\otimes$ represents the Kronecker product and $I_{i \times i}$ is the ${i \times i}$ identity matrix.
Now we can rewrite Equation~\ref{eq: redundantJacobian} as
\begin{equation}
    \dot{\boldsymbol{\theta}} = \boldsymbol{J_e}\boldsymbol{\dot{x}} + \boldsymbol{A_r} \dot{\boldsymbol{\theta_r}} = \boldsymbol{J_e}\boldsymbol{\dot{x}} + \boldsymbol{A_r} \mathcal{D}_{n-m} \boldsymbol{\theta_r}
\end{equation}
\begin{equation}
    \ddot{\boldsymbol{\theta}} = D_{n} \dot{\boldsymbol{\theta}}
\end{equation} 
where $\boldsymbol{J_e}$ and $\boldsymbol{A_r}$ are the stacked matrices formed from $J_e$ and $A_r$ respectively at each configuration.
At each timestep, the wrench is given by $h_{e,1},...h_{e,k}$, written as $\boldsymbol{h_e}$, and actuator efforts $u_{1},...u_{k}$, written as $\boldsymbol{u}$. Each timestep also has a unit vector $c_{1},...c_{k}$ along which the wrench should be maximised, written as  $\boldsymbol{c}$.
The aim is to solve for a set of redundant configurations along the trajectory which maximises the minimum wrench capability, a max-min optimisation. 

Given a set of configurations $\boldsymbol{\theta}$, the max-min wrench capability can be solved as a Linear Program (LP) over the entire trajectory, with objective
\begin{equation}
    \max \; t
    \label{eq: trajectoryLP}
\end{equation}
subject to 
\begin{equation}
  t  < \boldsymbol{c}^T \boldsymbol{h_e}
\end{equation}
where $t$ represents the minimum wrench capability. 
The equality constraints are given by the dynamics balance equations at each timestep $i$ are given by
\begin{equation}
    \tau_{d,i} + J_i^T h_{e,i} = Bu_i 
    \label{eq:constTraj1}
\end{equation}
Since the terms in $\tau_d$ are fully determined by $\theta,\dot{\theta}$ and $\ddot{\theta}$, the dynamic terms can be predetermined over the entire trajectory independently of $h_e$ and $u$.
There are inequality constraints
\begin{equation}
   \boldsymbol{u_{min}} \leq \boldsymbol{u} \leq \boldsymbol{u_{max}}
\end{equation}
where $\boldsymbol{u_{min}}$ and $\boldsymbol{u_{max}}$ are the stacked vectors of minimum and maximum actuator efforts respectively. The orthogonal wrench constraint is again given by
\begin{equation}
   (\boldsymbol{c}\boldsymbol{c}^+ - I_{6k})\boldsymbol{h_e} = 0
\end{equation} 
As before, this constraint can be relaxed under certain conditions. Finally there is a constraint on large changes to actuator efforts throughout the trajectory given by
\begin{equation}
    -\Delta  \boldsymbol{u_{min}} \Delta t \leq D_{mk}\boldsymbol{u} \leq \Delta \boldsymbol{u_{min}} \Delta t
    \label{eq:constTraj4}
\end{equation}
where $\Delta \boldsymbol{u_{min}}$ is the maximum allowed actuator change per second. This accounts for the relatively slow actuator dynamics of underwater thrusters.
The solution to this LP is written as $f(\boldsymbol{\theta_r})$ and is a function of the redundant configurations along the trajectory $\boldsymbol{\theta_r}$. Again the LP is solved using the dual-simplex method in Matlab's linprog function. The trajectory optimisation problem is given by
\begin{equation}
    \boldsymbol{\theta_r}_{opt} = \arg \min_{\boldsymbol{\theta_r}} -f(\boldsymbol{\theta_r})
    \label{eq:upperLevelTrajectory}
\end{equation}
where $\boldsymbol{\theta_r}_{opt}$ is the optimal set of redundant configurations along the trajectory. There are constraints on the configurations
\begin{equation}
    \boldsymbol{\theta_r}_{min} \leq \boldsymbol{\theta_r} \leq \boldsymbol{\theta_r}_{max}
\end{equation}
where $\boldsymbol{\theta_r}_{min}$ and $\boldsymbol{\theta_r}_{max}$ are the minimum and maximum limits on the configurations, and constraints on the velocities
\begin{equation}
    \boldsymbol{\dot{\theta_r}}_{min} \leq \boldsymbol{\dot{\theta_r}} \leq \boldsymbol{\dot{\theta_r}}_{max}
\end{equation}
where $\boldsymbol{\dot{\theta_r}}_{min}$ and $\boldsymbol{\dot{\theta_r}}_{max}$ are the minimum and maximum limits on the velocities.
This is again a bi-level optimisation problem, with the linear program in Equation~\ref{eq: trajectoryLP} as the lower level optimisation. Changes to $\boldsymbol{\theta_r}$ effect only the dynamics terms $M,h,g,\dot{\theta},\ddot{\theta}$ and the Jacobian $J_i$ in Equation~\ref{eq:constTraj1}. Therefore only the equality constraints are changed with changes in $\boldsymbol{\theta_r}$. Using the Karush-Kuhn-Tucker (KKT) conditions, the gradient of $f(\boldsymbol{\theta_r})$ with respect to $\boldsymbol{\theta_r}$ can be found as
\begin{equation}
    \frac{df(\boldsymbol{\theta_r})}{d\boldsymbol{\theta_r}} = \lambda_{eq}^T  \frac{dc_{eq}}{d\boldsymbol{\theta_r}}\Bigr|_{\substack{(u^*,h_e^*)}}
\end{equation}
where $\lambda_{eq}$ is the vector of Lagrange multipliers associated with the equality constraints, and the final term on the right is the gradient of the equality constraints $c_{eq}$, evaluated with the arguments of the solution to the lower level LP problem $(u^*,h_e^*)$. This is a non-convex objective with multiple local minima. An interior-point solver using the Matlab function fmincon is used, initiated at several different starting points in an attempt to find a good global minimum. 

Given an dynamically infeasible trajectory due to large dynamics terms, the lower level LP can not find a feasible solution which satisfies the constraints in Equation~\ref{eq:constTraj1}. In this case, the lower level problem is set to return a max-min wrench of  $f(\boldsymbol{\theta_r}) = \frac{1}{2}\tau_d^T\tau_d$, and gradient $\tau_d$ to the upper level problem, pushing the optimiser towards dynamically feasible trajectories.

\subsection{Tracking Dynamic Trajectories}
The above method finds a trajectory with smoothly varying actuator efforts, which maximises the minimum wrench in a given desired direction along the entire trajectory. The actual reaction forces and torques between the end effector and the environment during this trajectory will not be the same as those in the LP solution. 
Since the actuator constraints form a convex set, any value for $c^Th_e$ less than the maximum found by the optimisation will still fall within the feasible set. Additionally, assuming smooth changes in the interaction wrenches with the environment, and given the linear mapping between wrenches and each DOF, the smoothness properties are also conserved. The problem is finding the actuator efforts at a given timestep to track the desired trajectory, given that the actual efforts $\boldsymbol{u}$ may not match the LP solution.
At a given timestep $i$, the control effort $\tau_c$ to track the trajectory is computed using a appropriate controller~\cite{anisotropicDisturbance}. Given the overactuated system, the objective is to find the actuator efforts $u_c$ which are close to the optimal LP solution $u_i^*$, posed as a quadratic cost 
\begin{equation}
    \min_{u_c} (u_c - u_i^*)^T Q (u_c - u_i^*)
\end{equation}
with equality constraint
\begin{equation}
    B u_c = \tau_c
\end{equation}
which is simply the actuator model.
Using the method of Lagrange multipliers this has solution
\begin{equation}
    u_c^* = u_i + B_Q^+ (\tau_c - B u_i)
\end{equation}
where $B_Q^+$ is the weighted pseduoinverse given by
\begin{equation}
    B_Q^+ = Q^{-1}B^T(BQ^{-1}B^T)^{-1}
\end{equation}
This results does not incorporate the inequality constraints on the actuators. These are
\begin{equation}
    u_{min} \leq u_i \leq u_{max}
\end{equation}
which account for actuator effort limits and
\begin{equation}
    -\Delta  u_{min} \Delta t \leq D_{m}u_i \leq \Delta u_{min} \Delta t
\end{equation}
which account for actuator effort rate change limits. This is a quadratic programming problem, which can be solved efficiently in real time for control.

\section{Maximum Wrench Impulse}
\label{sec: Impulse}

This section looks at generating a maximum momentary wrench for a fixed end effector pose, using dynamic vehicle-manipulator motions while keeping the end effector fixed. A use case for this method may be for shifting a very heavy weight or a stuck valve. Humans naturally perform these motions for similar tasks, where the momentum of the person is used to momentarily generate large forces. 

\subsection{Problem Formulation}
For a given fixed end effector pose, and set of redundant configurations which define a dynamic trajectory, the maximum wrench problem is given by
\begin{equation}
    f = \max_{\boldsymbol{u},\boldsymbol{h_e}} \; \max(\boldsymbol{c}^T\boldsymbol{h_e})
\end{equation}
which is a linear $max-max$ problem. These problems can be reformulated as a Mixed Integer Linear Program (MILP) using the big-M method, although this is no longer a convex optimisation problem. Instead, the maximum wrench impulse can be chosen to occur at an arbitrarily chosen specific timestep $t_h$
\begin{equation}
    f = \max_{\boldsymbol{u},\boldsymbol{h_e}} \; \max(c_{t_h}^Th_{e,t_h})
\end{equation}
All constraints from the the dynamic wrench maximisation in Equations~\ref{eq:constTraj1}-~\ref{eq:constTraj4} also apply. This is a linear program which again is the lower level problem for the bi-level trajectory optimisation with upper level given by Equation~\ref{eq:upperLevelTrajectory}, with the solution and gradients found in the same way as before.

\section{Results}
\label{sec: Results}
We present the wrench maximisation results for the 4 DOF manipulator and 6DOF vehicle shown in Figure~\ref{fig:simulation}. This systems has a degree of redundancy of 4. Two wrench objectives are compared, torque along the z direction simulating turning a valve, and force in the z direction which tests lifting a heavy load. The torque limit on each joint is $[\pm 9, \pm 9, \pm 9, \pm 2]Nm$ and each thruster has an asymmetrical thrust limit of -40N to 50N. The force of each thruster can vary by 50N/s, while the rate of change of joint torque is effectively instantaneuous. Six load cells were attached in series shown in Figure~\ref{fig:simulation}, allowing all six components of the applied wrench to be measured simultaneously. The sensor setup was calibrated using known masses and levers.

\subsection{Static Wrench Capability}
\begin{figure} [t]
   \centering
   \includegraphics[width=1\linewidth]{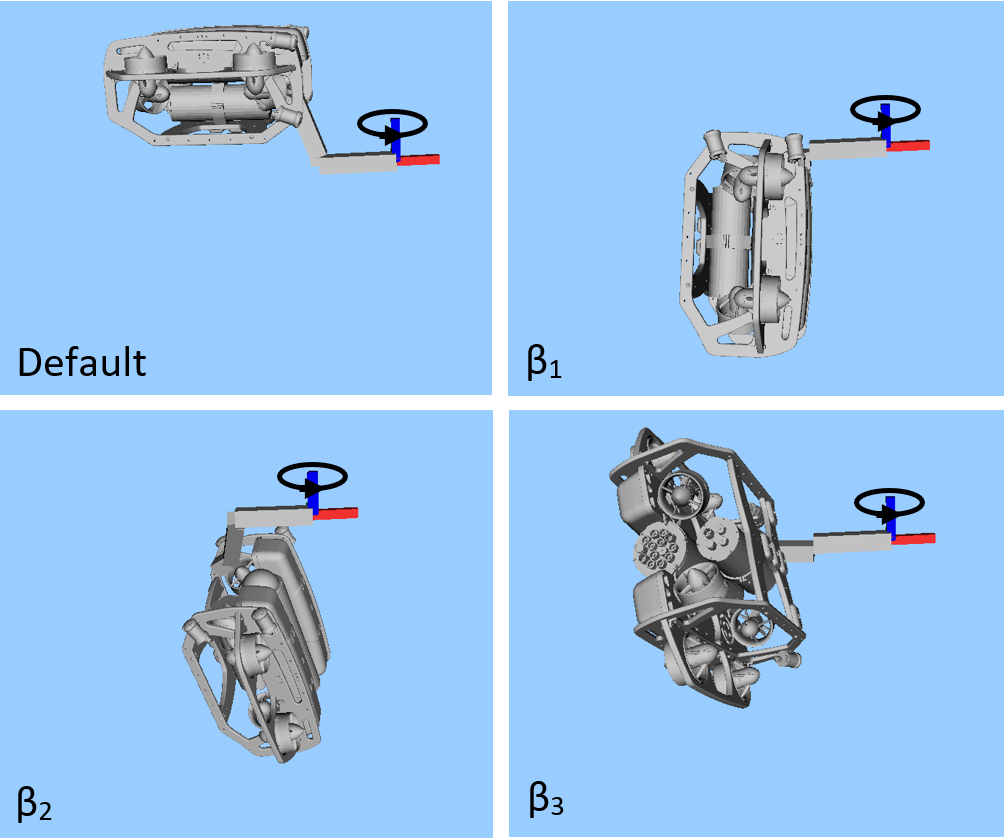}
   \caption{Diagrams showing optimal static configuration for torque in the z direction, comparing the default upright case, to the solutions which optimise $\beta_1$, $\beta_2$ and $\beta_3$}
   \label{fig:optimalConfigurations}
\end{figure}

For the case of static torque maximisation in the z direction, four configurations are compared, shown in Figure~\ref{fig:optimalConfigurations}. These  first is the default configuration, which is simply the manipulator in a neutral position with the vehicle upright. The other three cases are the optimised configurations for $\beta_1, \beta_2$ and $\beta_3$ as described in section~\ref{sec: Static Wrench}, with the optimal thruster forces and manipulator torques computed accordingly. 

Figure~\ref{fig: staticTorque} shows the torque along the z direction as measured by the sensor setup during the experiments for each of the four configurations. Each experiment consists of a 5 seconds ramp-up and ramp-down and a 5 second maximum wrench. 
\begin{figure} [t]
   \centering
   \includegraphics[width=1\linewidth]{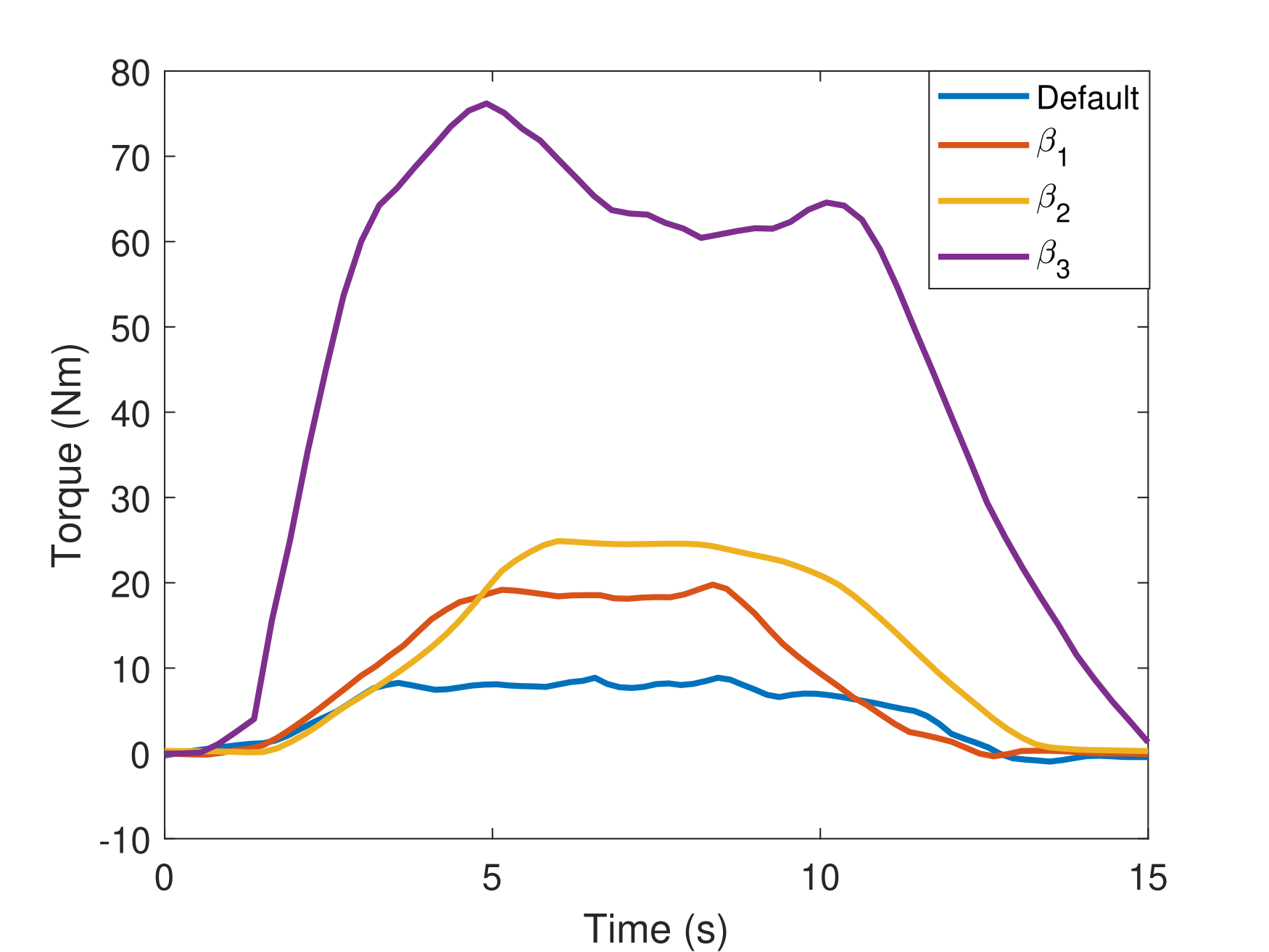}
   \caption{Torque along the z direction during static wrench maximisation, comparing the default, $\beta_1$, $\beta_2$ and $\beta_3$}
   \label{fig:staticTorque}
\end{figure}

The default case is limited to a torque of 9Nm, since the z axis is aligned with the base joint of the manipulator in this case. Optimising for the transmission ratio $\beta_1$ results in a maximum torque of approximately 19.5Nm, while optimising using the maximum wrench polytope $\beta_2$ gives a maximum of 25Nm. Optimising for the maximum wrench polytope with the constraint on orthogonal wrench components removed leads to a maximum torque of 76Nm. The inconsistent torque readings during the fixed maximum thrust period are due to the effects of swirling water after several seconds in the relatively small test tank. 

\begin{figure} [t]
   \centering
   \includegraphics[width=1\linewidth]{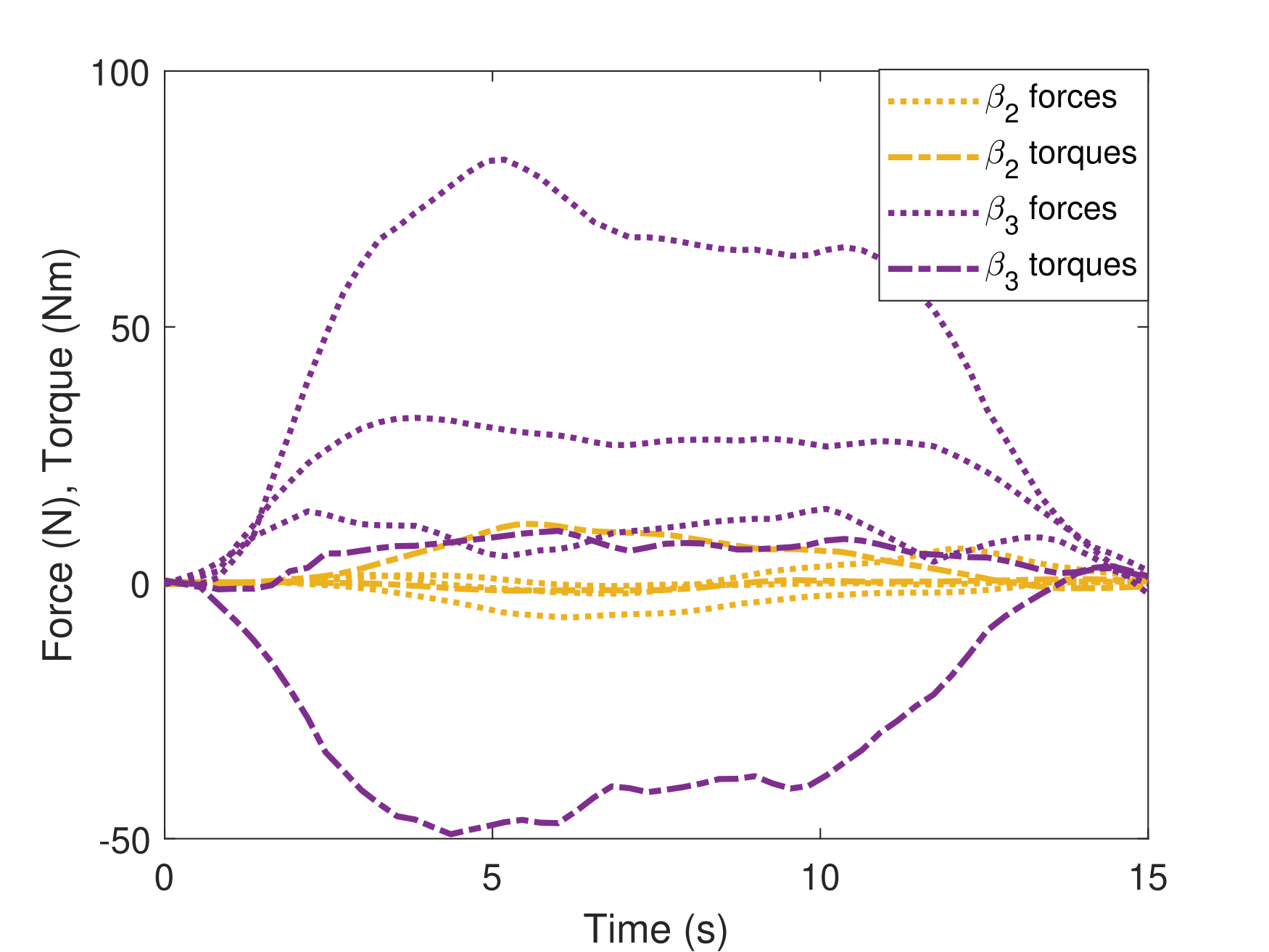}
   \caption{Forces in xyz and torques in xy during static wrench maximisation in the z direction, comparing the default, $\beta_2$ and $\beta_3$}
   \label{fig:orthogonal}
\end{figure}

The default, $\beta_1$ and $\beta_2$ cases all have constraints on wrenches orthogonal to torque along the z direction. The $\beta_3$ case relaxes this constraint and has significant orthogonal components. Figure~\ref{fig:orthogonal} shows each of the 3 orthogonal forces and two orthogonal torques during the the experiment. All the orthogonal components during the $\beta_2$ experiment are relatively small, with a peak of around 10Nm along the y direction due to flexible strain in the sensor and UVMS setups. In contrast, the $\beta_3$ experiment has significant forces of around 80N and torque of around 50Nm. These may be acceptable in a case such as turning a valve, where the valve can resist these orthogonal wrenches.

\begin{figure} [t]
   \centering
   \includegraphics[width=1\linewidth]{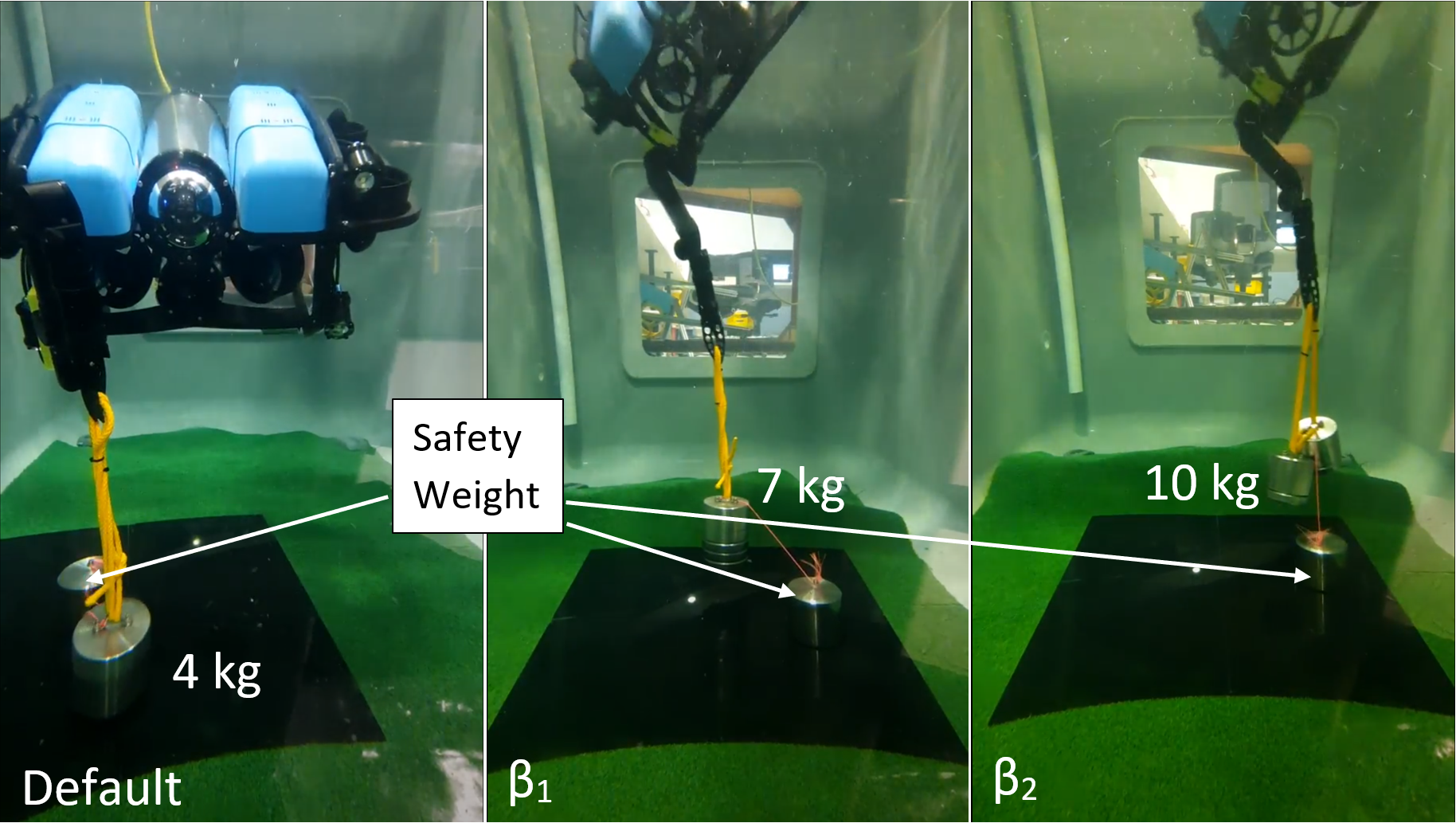}
   \caption{Experiments comparing comparing the default, $\beta_1$ and $\beta_2$ force maximisation cases, showing the lifted mass. Each test mass is attached to a safety weight to keep the system constrained.}
   \label{fig:staticForce}
\end{figure}

For the case of static force maximisation in the z direction, three configurations are compared, shown in Figure~\ref{fig:staticForce}. Again the first is the default configuration, followed by optimal configurations for $\beta_1$ and $\beta_2$. In this case testing for $\beta_3$ is not valid, as the weight should lifting vertically up and not accelerated in orthogonal directions. This test was performed using known weights, and the successfully lifted masses are shown in Figure~\ref{fig:staticForce}. The default case can lift a maximum of 4kg, the optimisation using $\beta_1$ can lift 7kg, and the optimisation using $\beta_2$ can lift 10kg.

The results for both the torque maximisation and force maximisation experiments shows significant increases of approximately 30\% and 40\% respectively in wrench capability when comparing optimisation using the transmission ratio $\beta_1$, as compared to the proposed method of optimising the directional wrench polytope using a bi-level optimisation with a LP lower level problem $\beta_2$. Additionally, in the case when the orthogonal wrench constraints can be removed, the experimental torque capability increased threefold. 

\begin{figure} [h!]
   \centering
   \includegraphics[width=1.0\linewidth]{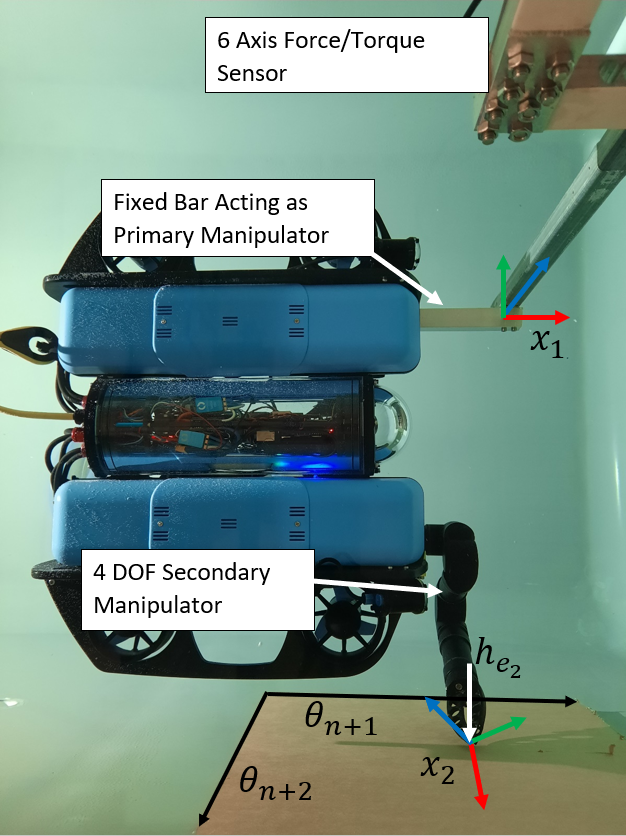}
   \caption{Experimental setup for multiple contacts points. A fixed bar is used as the primary manipulator, with end effector pose $x_1$, and a 4DOF manipulator is used as the secondary manipulator with end effector pose $x_2$. The secondary manipulator is contacting the side of the tank where a normal force into the wall can be applied, labelled as $h_{e_2}$. Possible regrasping points for the second manipulator which provide additional DOF over which to optimise are labelled as $\theta_{n+1}$ and $\theta_{n+2}$}
   \label{fig:dualSetup}
\end{figure}

\subsection{Multiple Contact Points}
The experimental setup for multiple contact point is shown in Figure~\ref{fig:dualSetup}.  A fixed bar is used as the primary manipulator with end effector $x_1$, and is attached to the 6 axis force-torque sensor setup. A 4DOF manipulator is used as the secondary manipulator with end effector pose $x_2$, which is shown in contact with the side of the tank where a normal force can be applied. The first constraint in~\ref{eq: dualOrthogonal} is also applied, limiting the wrench applied at $x_1$ to a pure torque in the z direction. It is assumed all forces and torques at $x_2$ are 0 except for a positive normal force into the wall. Therefore $C_2$ only has one component and the second constraint in~\ref{eq: dualOrthogonal} applies, as well as in inequality constraint in Equation~\ref{eq: inequalityContact} to ensure the contact force is into the wall. 


\begin{figure} [h!]
   \centering
   \includegraphics[width=0.9\linewidth]{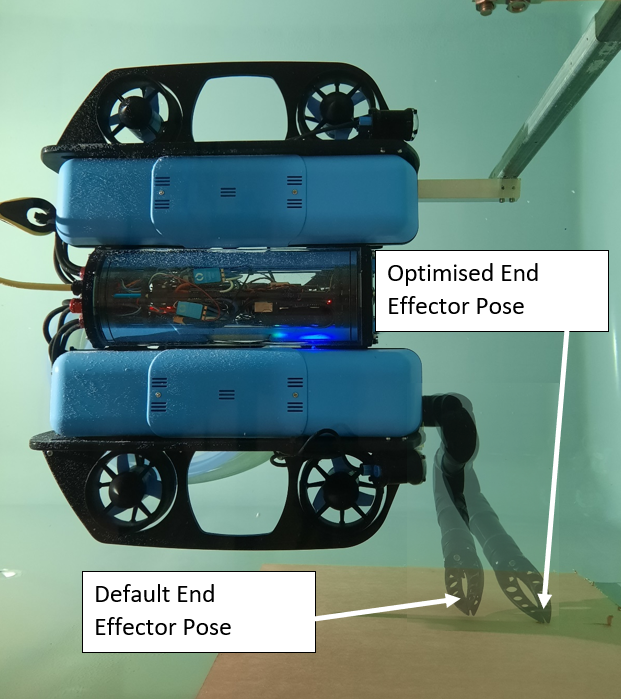}
   \caption{Default and optimised configurations and end effector poses for a secondary contact point during torque maximisation}
   \label{fig:dualOptimal}
\end{figure}

\begin{figure} [h!]
   \centering
   \includegraphics[width=1\linewidth]{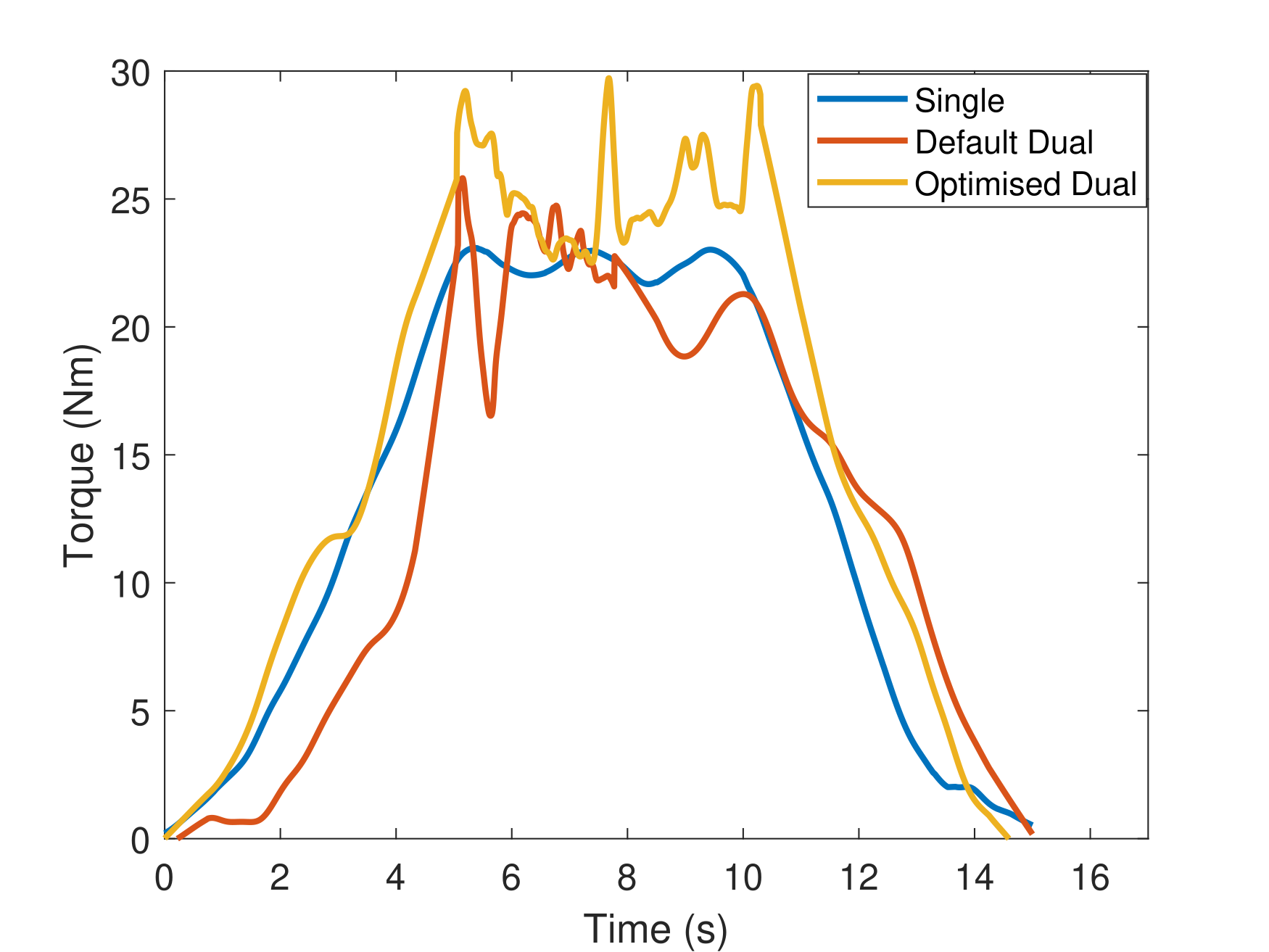}
   \caption{Torque along the z direction during static wrench maximisation comparing single manipulator, dual manipulators in the default end effector pose, and dual manipulators in the optimised end effector pose}
   \label{fig:dualTorque}
\end{figure}

Figure~\ref{fig:dualSetup} shows the set of possible regrasping points which provide two additional DOF over which to optimise. Figure~\ref{fig:dualTorque} shows the results for torque maximisation along the z direction, comparing the single and dual manipulator cases. For the dual manipulator case, both the default end effector pose and the optimised end effector pose are tested, shown in Figure~\ref{fig:dualOptimal}. The single manipulator torque reaches a maximum of 23Nm, the default dual case reaches 26Nm, and the optimised dual case reaches 30Nm, an increase of 13\% and 30\% respectively. The data for this experiment contains more noise since the vehicle is very close to the edge of the tank by necessity, causing large effects from swirling water in the tank. These effects may be significant near solid structures in underwater environment, requiring additional modelling which is left as future work.

\subsection{Wrench Maximisation over a Trajectory}

\begin{figure} [b]
   \centering
   \includegraphics[width=1.0\linewidth]{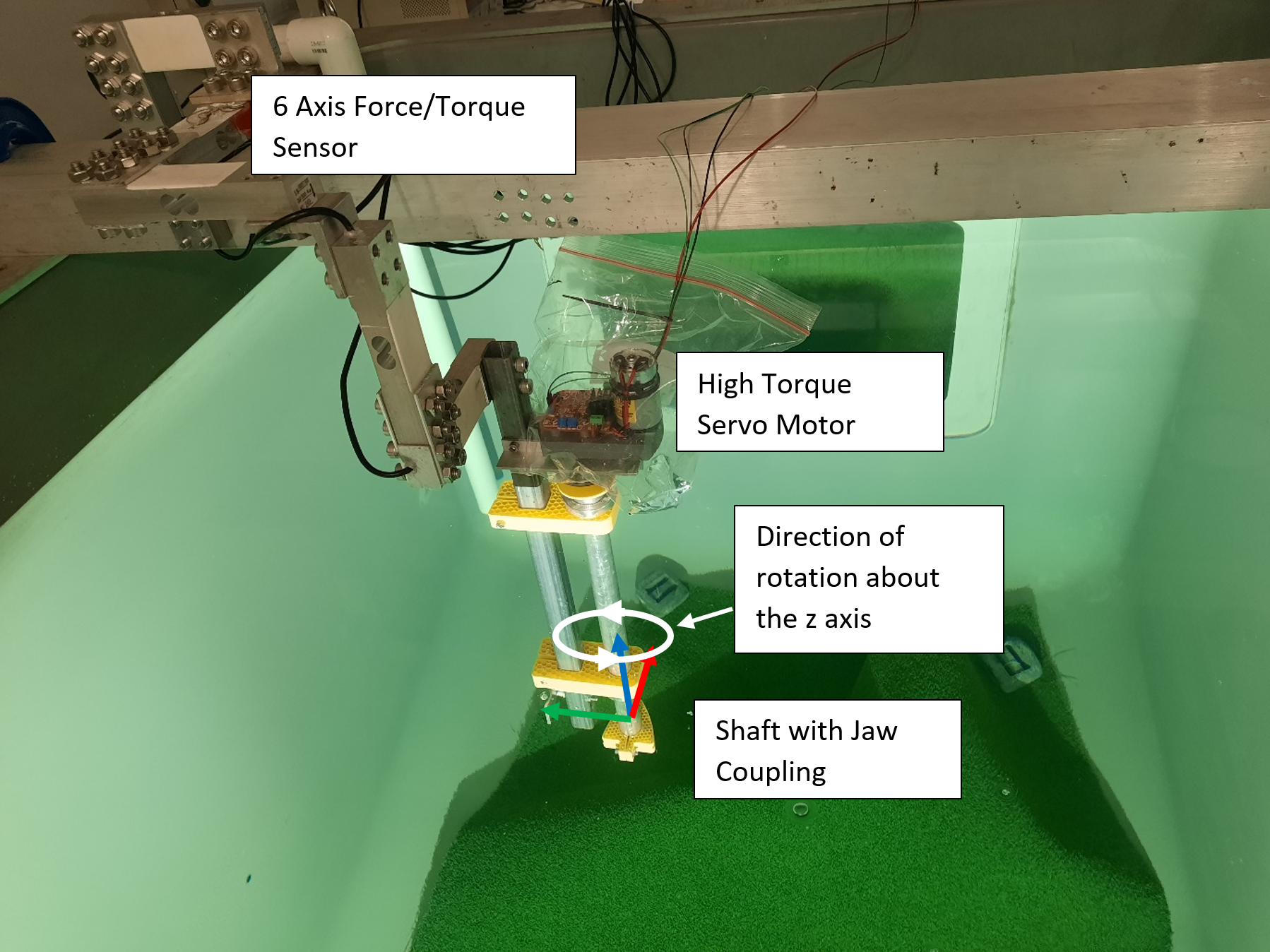}
   \caption{Setup for testing dynamics wrenches over a trajectory, suing a servo motor and attached shaft to simulate a turning valve}
   \label{fig:trajectorySetup}
\end{figure}

The experimental setup for testing wrenches over a rotating end effector trajectory along the global z axis is shown in Figure~\ref{fig:trajectorySetup}. A high torque servo motor above the water line is used to generate the controlled rotation, and is attached to a shaft which reaches into the tank. The end effector of the UVMS attaches to a jaw coupling on the end of the shaft, and the whole setup is mounted to the end of the 6 axis force-torque sensor. 

\begin{figure} [t]
   \centering
   \includegraphics[width=1.0\linewidth]{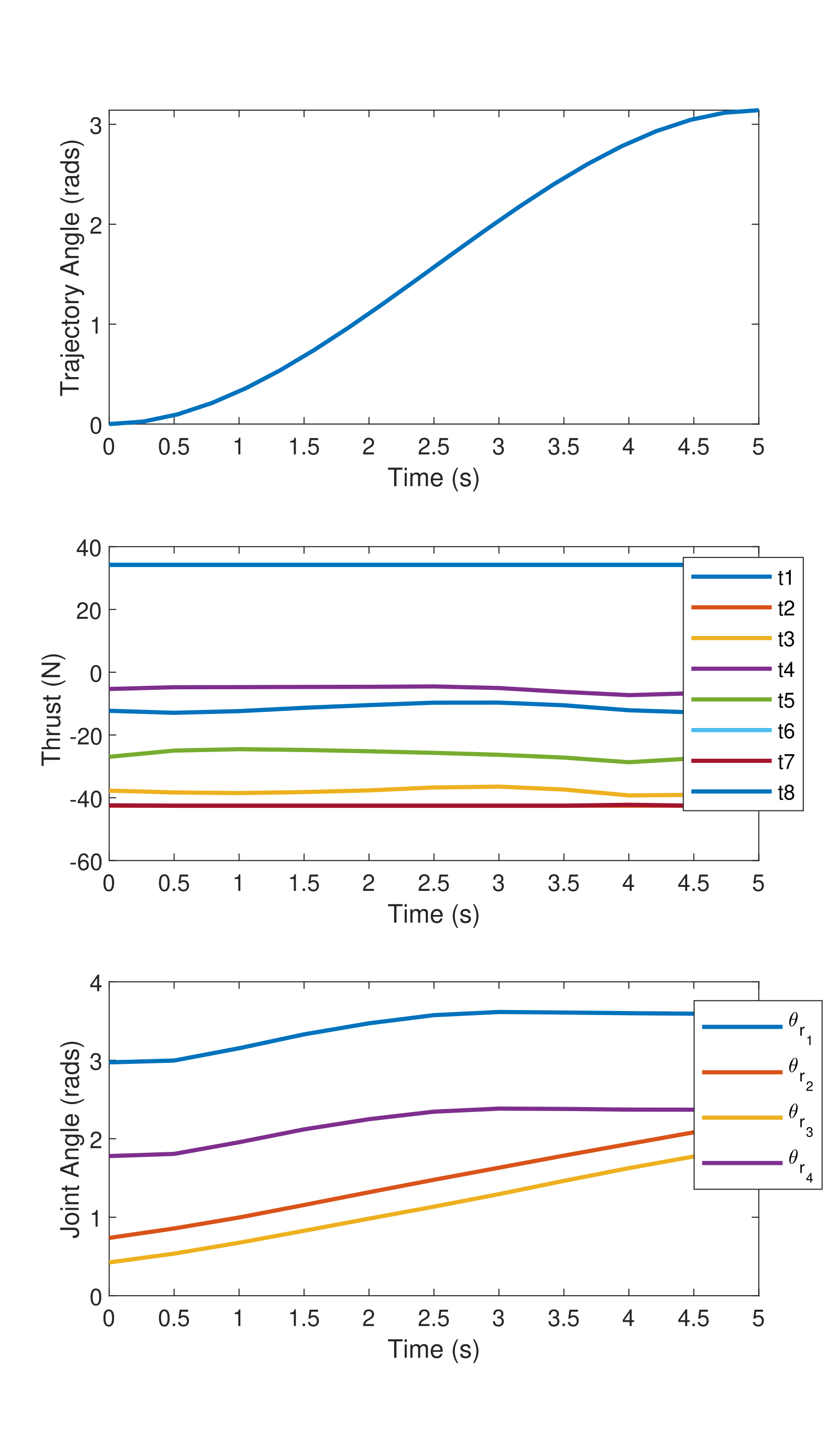}
   \caption{Plots during the rotating end effector trajectory showing (top) the end effector angle, (middle) vehicle thrust forces, and (bottom) redundant joint angles}
   \label{fig:optimisationTrajectory}
\end{figure}

The objective which is tested is maximising the torque along the z axis with no orthogonal wrench components, while the end effector tracks a $180^{\circ}$ rotating trajectory. The trajectory is shown in the top plot of Figure~\ref{fig:optimisationTrajectory}, and is generated as a cubic spline with a velocity of 0 at each end point, with a total time of 5 seconds. 10 points along the trajectory are considered, therefore $T=10$ for the trajectory optimisation problem described in Section~\ref{sec: Trajectory}. The values for each joint angle and thruster force were interpolated linearly between each timestep when sending commands to the UVMS during the experiment. The middle plot in Figure~\ref{fig:optimisationTrajectory} shows the thruster forces throughout the trajectory, showing the effect of the constraint in Equation~\ref{eq:constTraj4} which limits large changes in actuator effort. Finally the bottom plot in Figure~\ref{fig:optimisationTrajectory} shows the joint angles which correspond to the redundant DOF, showing a relatively smooth change in redundant configuration throughout the rotation.

\begin{figure} [t]
   \centering
   \includegraphics[width=1.0\linewidth]{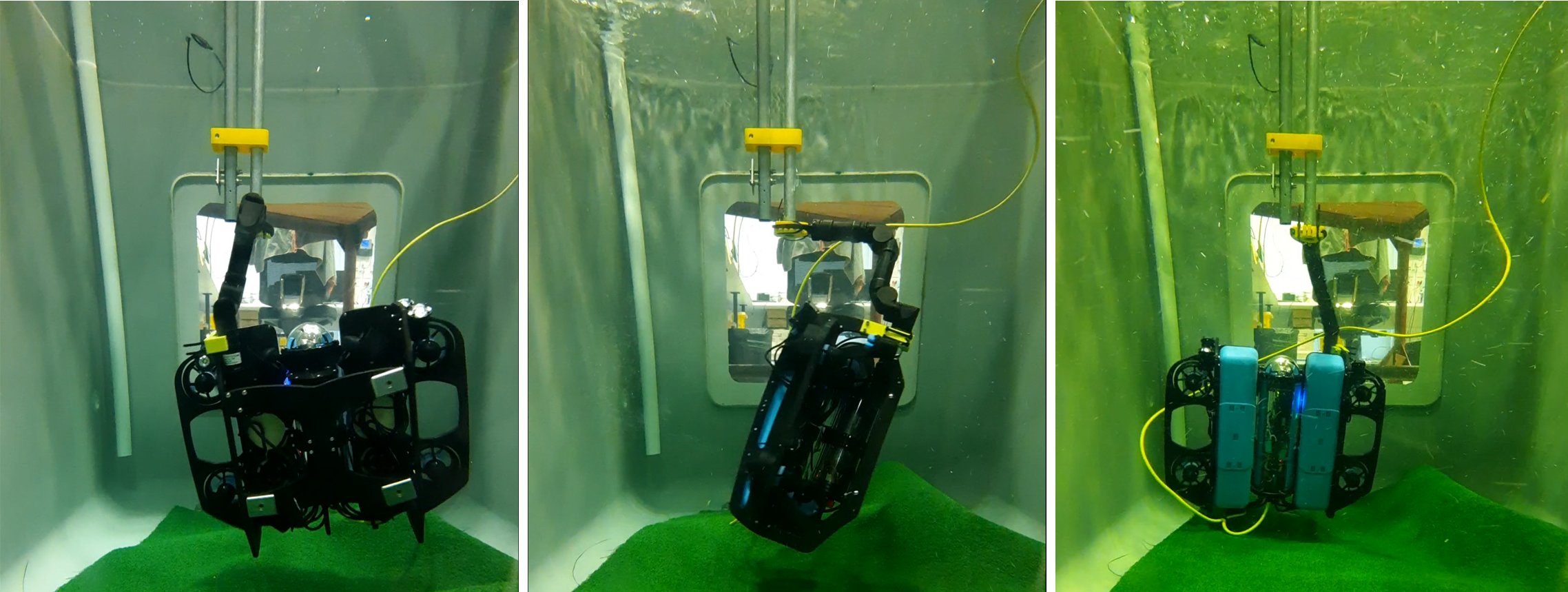}
   \caption{A sequence of images during the 5 second $180^{\circ}$ rotating trajectory, showing the changing manipulator joint angles during the rotation. The images are at $0^{\circ}$, $90^{\circ}$ and $180^{\circ}$}
   \label{fig:trajectorySequence}
\end{figure}

Figure~\ref{fig:trajectorySequence} shows a sequence of images of the UVMS during the experiment at at $0^{\circ}$, $90^{\circ}$ and $180^{\circ}$, showing the changing joint angles throughtout.

\begin{figure} [t]
   \centering
   \includegraphics[width=1\linewidth]{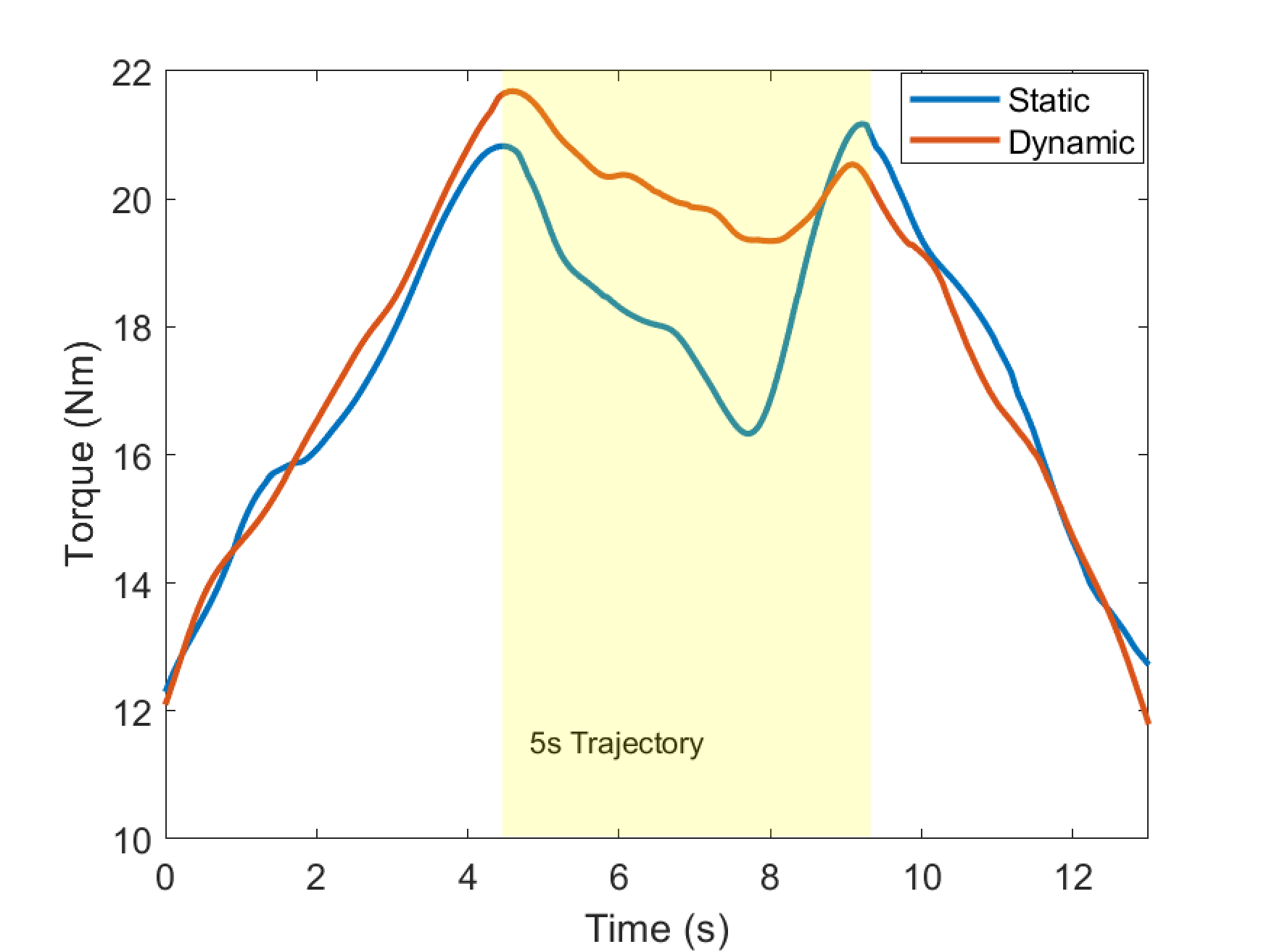}
   \caption{Torque along the z direction during a dynamic rotating end effector trajectory, comparison the static and dynamic redundant configurations cases}
   \label{fig:trajectoryResults}
\end{figure}

Figure~\ref{fig:trajectoryResults} shows the results of the torque along the z axis, comparing the dynamic case which tracks the optimised trajectory, and the static case which maintains a fixed redundant configuration throughout. The static redundant configuration which is chosen is the configuration which maximises the static torque capability using $\beta_2$. The  static case achieves a minimum torque of 16Nm throughout the 5 second rotation period, while the dynamic trajectory achieves a minimum torque of 19Nm, a 19\% increase.
The results show that consideration of the dynamical effects during the rotating end effector trajectory leads to a significant increase in the max-min torque. 

Since this rotation if about the vertical axis, the effects of changing gravity and buoyancy vectors in the UVMS frame due to the end effector rotation are not present. Therefore the improved performance of the dynamic case as compared to the static case is purely due to consideration of dynamical effects from velocity and acceleration. The difference in performance between the dynamic trajectory optimised case and the static case are likely to be much more significant with end effector trajectories which include rotations about the non-vertical axis.

\subsection{Maximum Wrench Impulse}

\begin{figure} [t]
   \centering
   \includegraphics[width=1.0\linewidth]{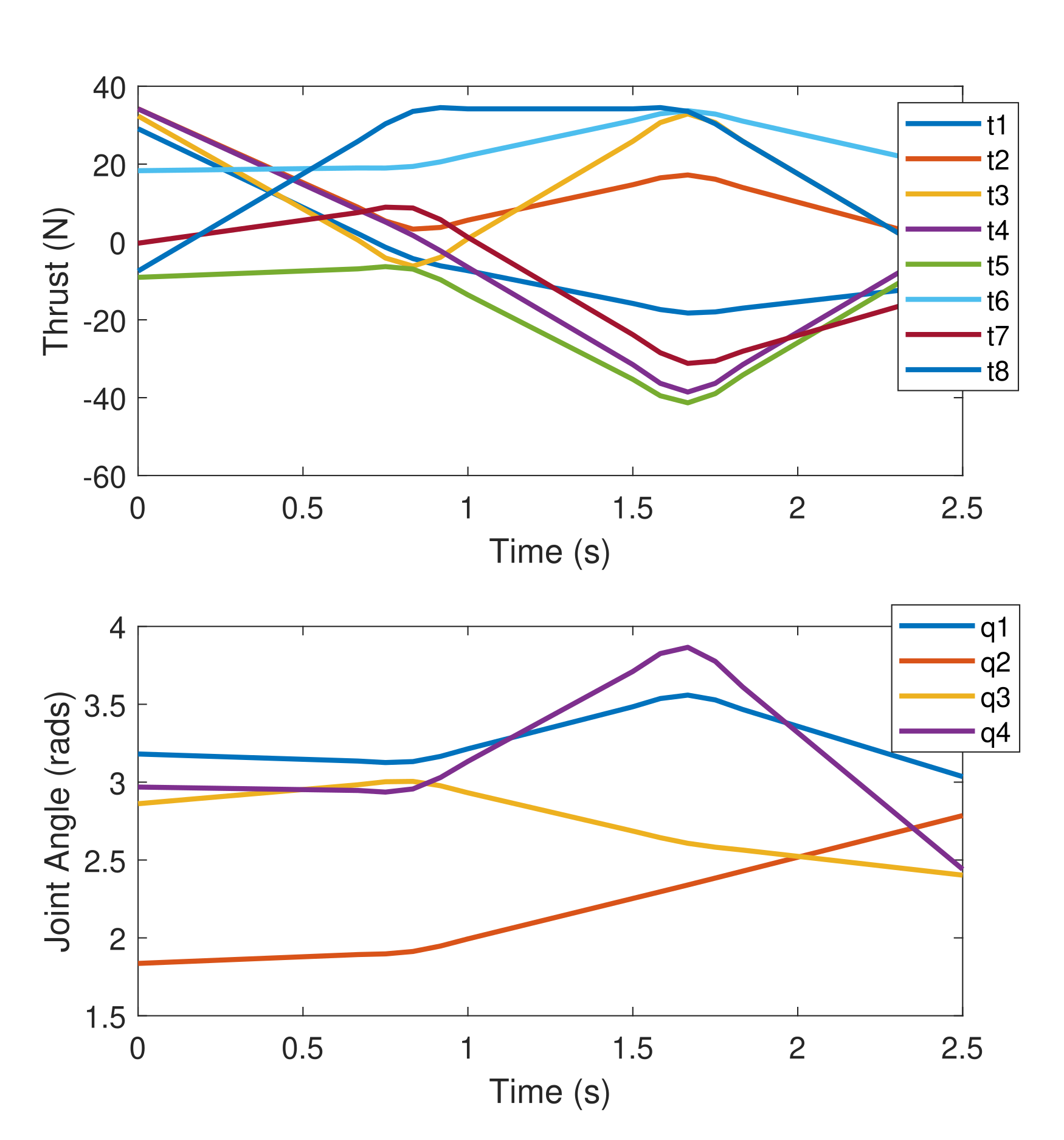}
   \caption{Plots during the maximum wrench impulse trajectory showing (top) the vehicle thrust forces, and (bottom) redundant joint angles}
   \label{fig:optimisationImpulse}
\end{figure}

\begin{figure} [b]
   \centering
   \includegraphics[width=1.0\linewidth]{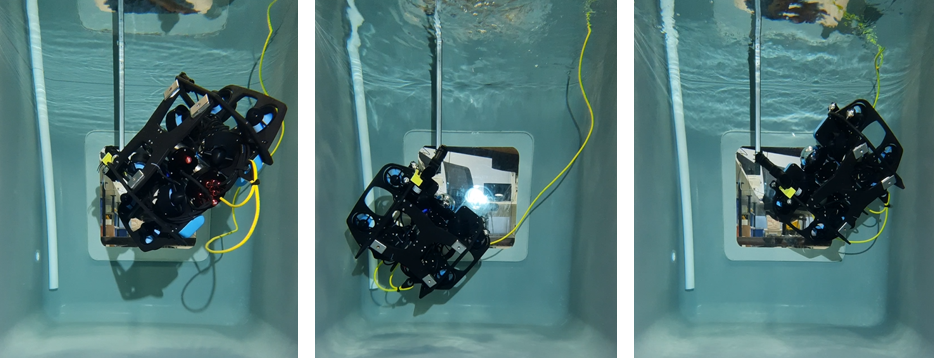}
   \caption{A sequence of images during the 3 second maximum impulse trajectory}
   \label{fig:impulseSequence}
\end{figure}

\begin{figure} [t]
   \centering
   \includegraphics[width=1\linewidth]{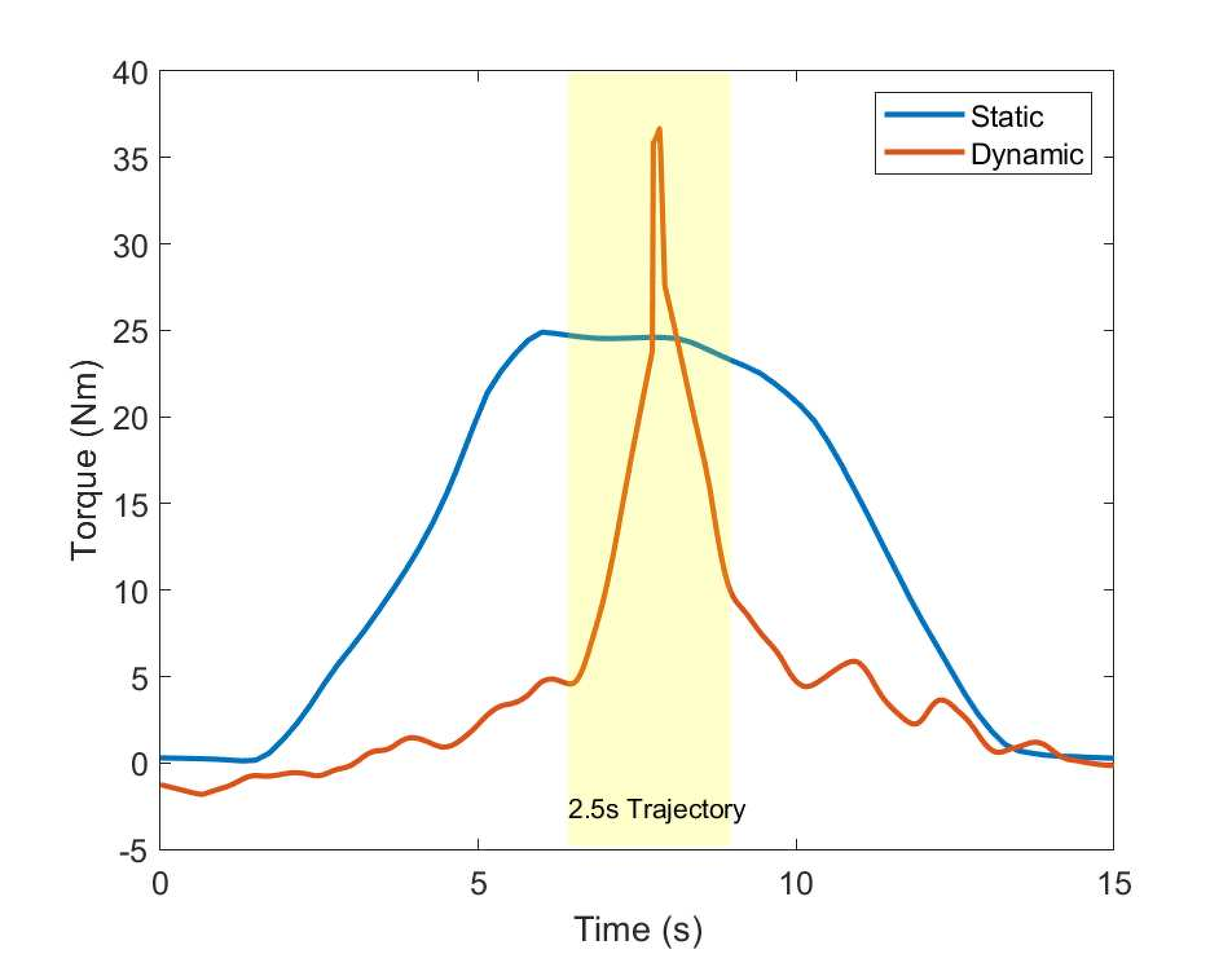}
   \caption{Torque along the z direction during a heaving motion, compared to the static $\beta_2$ case}
   \label{fig:impulseResults}
\end{figure}

The experimental setup for maximum wrench impulse trajectories is the same as the static wrench maximisation setup shown in Figure~\ref{fig:simulation}. Again the objective is to maximise the torque along the z direction with no orthogonal wrench components, in this case by using dynamic motions to generate a large torque impulse momentarily, while keeping the end effector fixed. A trajectory of 2.5s with 4 intermediate configurations was chosen, therefore $T=4$ for the trajectory optimisation problem described in Section~\ref{sec: Impulse}. The 3rd timestep is chosen as the point of maximum torque, therefore $t_h = 3$. As before, values for each joint angle and thruster force are interpolated linearly between each timestep when sending commands to the UVMS during the experiment.
Figure~\ref{fig:optimisationImpulse} shows the results of the trajectory optimisation, with the top plot showing the thruster forces throughout the trajectory, and the bottom plot shows the joint angles. The resulting trajectory leads to both joint velocities and thruster rates of change at the limits set in the optimisation problem, leading to a highly dynamic trajectory. Figure~\ref{fig:impulseSequence} shows a sequence of images during the trajectory, and Figure~\ref{fig:impulseResults} shows a plot of the measured torque results. The plot compares the static case which is identical to the $\beta_2$ optimised results from Figure~\ref{fig:staticTorque}, compared to the dynamic impulse case. The dynamic trajectory generated a torque impulse of 35Nm, which is a 40\% torque improvement over the best static results with orthogonal wrench constraints.

\section{Conclusion}\label{sec: Conclusions}
This work looks at maximising the maximum wrench capability of UVMS. A bi-level optimisation method is proposed for maximising static wrenches, and experimental results show a significant improvement over optimising the transmission ratio. Further results show that relaxing constrains on orthogonal wrenches leads to significant increases in wrench capability in relevant use cases. The case of multiple contact points, as well as re-grasping of secondary points is also considered, with experimental results again showing increased wrench capability. A similar bi-level optimisation approach is introduced for max-min wrench optimisation over a trajectory, with experimental results confirming the validity of the method. Finally, a method is proposed for finding dynamic trajectories which generate large wrench impulses, with supporting experimental results. Further work is required for dealing with the effects of self-generated currents by the vehicle thrusters when operating near underwater structures. Additional work would look at automatic recognition of viable secondary contact points.


\addtolength{\textheight}{-5cm}   





\section*{ACKNOWLEDGMENT}
This work has been enabled by use of a Reach Alpha manipulator from Blueprint Lab (https://blueprintlab.com/)

\bibliographystyle{IEEEtran}

\bibliography{bibtex/references}

\end{document}